%% file: main.tex
\documentclass{article}

\usepackage[final]{neurips_2025}

\usepackage{multirow}
\usepackage{wrapfig}
\usepackage[utf8]{inputenc} % allow utf-8 input
\usepackage[T1]{fontenc}    % use 8-bit T1 fonts
\usepackage{hyperref}       % hyperlinks
\usepackage{url}            % simple URL typesetting
\usepackage{booktabs}       % professional-quality tables
\usepackage{amsfonts}       % blackboard math symbols
\usepackage{nicefrac}       % compact symbols for 1/2, etc.
\usepackage{microtype}      % microtypography
\usepackage{xcolor}         % colors
\usepackage{amsmath}
\usepackage{xfrac}
\usepackage[capitalize, noabbrev]{cleveref}
\usepackage{enumitem}    
\usepackage{color,soul}  % for highlighting
\usepackage{caption}
\usepackage{footmisc}

\newcommand{\tpms}[1]{% small stddev, with plus-minus
\raisebox{0.0ex}{\scriptsize{\ensuremath{\,\pm\,}#1}}%
} 
\newcommand{\nuclr}{\texttt{NuCLR}} % us!

\newcommand{\noreffootnote}[1]{
    \begingroup
    \renewcommand{\thefootnote}{}
    \footnotetext{#1}
    \addtocounter{footnote}{0}
    \endgroup
}

\newcommand{\codelink}{\href{https://github.com/nerdslab/nuclr}{\texttt{https://github.com/nerdslab/nuclr}}}

\title{Know Thyself by Knowing Others: Learning Neuron Identity from Population Context}

\author{
{\bf Vinam Arora$^{1\dagger}$,
Divyansha Lachi$^{1}$,
Ian J. Knight$^{1}$,
Mehdi Azabou$^{2}$,} \\
{\bf 
Blake Richards$^{3,4}$,
Cole Hurwitz$^{2}$,
Joshua H. Siegle$^{5}$,
Eva L. Dyer$^{1\dagger}$} \\[1mm]
$^{1}$University of Pennsylvania,
$^{2}$Columbia University, \\
$^{3}$McGill University,
$^{4}$Mila,
$^{5}$Allen Institute for Neural Dynamics
}

\begin{document}
\maketitle
\captionsetup{font=footnotesize}

\noreffootnote{
$^{\dagger}$Contact: vinam@upenn.edu, dyer1@upenn.edu 
\\
\hspace*{2.2em}
Reviewed on OpenReview:
\href{https://openreview.net/forum?id=zt3RKc6VBp}{\texttt{https://openreview.net/forum?id=zt3RKc6VBp}}.
}

\input{files/abstract}
\input{files/intro}

\input{files/methods}

\input{files/results}

\input{files/relatedworks}
\input{files/discussion}
\input{files/funding}

\bibliography{main.bib}
\bibliographystyle{abbrv}

\newpage
\input{appendix/main}

\end{document}

%% file: files/abstract.tex
\begin{abstract}
Neurons process information in ways that depend on their cell type, connectivity, and the brain region in which they are embedded.
However, inferring these factors from neural activity remains a significant challenge.
To build general-purpose representations that allow for resolving information about a neuron's identity, we introduce \texttt{NuCLR}, a self-supervised framework that aims to learn representations of neural activity that allow for differentiating one neuron from the rest. 
\texttt{NuCLR} brings together views of the same neuron observed at different times and across different stimuli and uses a contrastive objective to pull these representations together. 
To capture population context without assuming any fixed neuron ordering, we build a spatiotemporal transformer that integrates activity in a permutation-equivariant manner.
Across multiple electrophysiology and calcium imaging datasets, a linear decoding evaluation on top of \texttt{NuCLR} representations achieves a new state-of-the-art for both cell type and brain region decoding tasks, and demonstrates strong zero-shot generalization to unseen animals.
We present the first systematic scaling analysis for neuron-level representation learning, showing that increasing the number of animals used during pretraining consistently improves downstream performance. The learned representations are also label-efficient, requiring only a small fraction of labeled samples to achieve competitive performance. These results highlight how large, diverse neural datasets enable models to recover information about neuron identity that generalize across animals. 
Code is available at ~\codelink.
\end{abstract}

%% file: files/intro.tex
\section{Introduction}
Neurons operate in ways that reflect their cell type, connectivity, and the brain region in which they are embedded \cite{zeng2017neuronal,tolossa2024conserved,schneider2025robust}. Understanding how these structural and physiological factors shape neural activity is an important open question \cite{schneider2025robust} and is central to modern theories of neural function \cite{jha2024disentangling, christensen2022cognition}. 
Cell type- and circuit-specific effects are also increasingly implicated in neurological disorders, where it has been shown that particular neuronal subtypes can be selectively affected in neurodegenerative disease  \cite{pak2024distinctive, mcgregor2024failure} and that neuron subtypes can be targeted to improve efficacy in real time in closed-loop brain stimulation \citep{zhao2025distinct}. 
Together, these directions highlight a broader need to understand how features such as cell type and brain region contribute to a neuron's role in the brain, motivating the development of methods that can extract aspects of neuron identity.

Much of what we know about neuronal cell types and their connectivity comes from molecular, anatomical, and morphological labeling techniques that tag neurons based on gene expression profiles or structural features \cite{zeng2017neuronal,tasic2018shared,gouwens2019classification,schneider2025inhibitory}.
These approaches have revealed a rich taxonomy of neuronal classes, but they 
are resource intensive and typically provide only a sparse labeling of one or a few classes at a time. 
Currently, there is no molecular access to different cell types or projection classes in nonhuman primates and humans  \cite{juavinett2019chronically, mosher2020cellular, lee2024large}, which has created major roadblocks in studying neuron diversity in humans. For these reasons, there is a growing interest in methods that can infer aspects of neuronal identity directly from large-scale neural population activity.
 
Recent work has started to explore deep learning approaches for tackling this question \cite{schneider2023transcriptomic, mi2023learning, tolossa2024conserved, beau2025deep, yu2025in, wu2025neuron}. 
Early methods relied on intrinsic single-neuron features such as waveform shape or interspike interval profiles \cite{schneider2023transcriptomic, tolossa2024conserved, beau2025deep, yu2025in}, but these signals capture only part of what distinguishes one neuron from another. 
More recent approaches have attempted to incorporate some information about the ongoing population activity \cite{mi2023learning, wu2025neuron}, but  reduce the surrounding activity to a fixed low-dimensional summaries, which obscures the fine-grained coordination patterns that are often key to neuronal identity. 
Furthermore, these models usually require retraining or finetuning on each new set of neurons, limiting their ability to generalize in a zero-shot manner across animals. These limitations point to the need for methods that leverage rich population context while producing neuron-level representations that transfer robustly across recordings. 

To fill this gap, we introduce \texttt{NuCLR}, a self-supervised approach for learning neuron-level representations from large-scale, neural activity datasets.
At a high-level, the goal of \texttt{NuCLR} is to build a representation that is stable across time and makes each neuron distinguishable from other neurons in the population, thereby capturing abstract features that reflect its identity.
Within these representations, we expect to uncover information related to cell type, brain region, connectivity, and potentially other latent attributes. To learn these representations in a self-supervised manner, 
\texttt{NuCLR} constructs two population-level views by sampling and encoding different temporal segments from the same recording, and applies a contrastive objective that treats representations of the same neuron across these views as positives and all other neurons in the population as negatives. 
To allow for integration of rich population context in each neuron's representation, we introduce a spatiotemporal transformer that aggregates information from surrounding neurons without assuming a fixed ordering or requiring session-specific alignment.
Together, these components allow \texttt{NuCLR} to learn population-aware representations that generalize reliably across animals, enabling zero-shot decoding of neuron-level attributes from previously unseen recordings.

To evaluate the identity-related information learned by \texttt{NuCLR}, we first pretrain the model in an unsupervised manner and then train linear classifiers on the resulting representations to decode cell type and brain region. 
Across multiple electrophysiology and calcium imaging datasets, \texttt{NuCLR} achieves a new state-of-the-art on both tasks.
We show that \nuclr's population-based self-supervision yields representations that transfer robustly across sessions and animals in a \textit{zero-shot} manner, supporting decoding neuron identities on entirely new subjects without retraining or additional metadata.
In addition to strong zero-shot generalization, \nuclr~is highly label efficient: using only 12.5\% of labels to train the classifier still outperforms all baselines even when they use all labeled data.
We further provide the first systematic scaling analysis for neuron-level tasks, demonstrating that increasing the number of  animals used in pretraining consistently improves zero-shot cell type and region decoding accuracy. On the Allen Visual Coding Neuropixels dataset \cite{siegle2021survey}, doubling the unlabeled pretraining data improves cell-type decoding more than doubling the supervised labels, highlighting the advantages of scaling pretraining across many animals.

\vspace{2mm}
\textbf{Our core contributions are:}
\begin{itemize}[leftmargin=5mm,itemsep=1mm,topsep=1mm]

\item We introduce \texttt{NuCLR}, a self-supervised framework for learning neuron-level representations from neural population activity. After unsupervised pretraining, these representations support linear decoding of cell type and brain region and set a new state-of-the-art across multiple electrophysiology and calcium imaging datasets.

\item We show that \texttt{NuCLR} generalizes in a {\em zero-shot} manner: the same pretrained model and embeddings transfer robustly to entirely new sessions and animals without retraining or requiring additional metadata, enabling out-of-the-box decoding of cell type and brain region.

\item We provide the first systematic scaling analysis for neuron-level representation learning, demonstrating that increasing the number of animals used during pretraining yields consistent gains in zero-shot cell-type and region decoding. This underscores the value of large, diverse unlabeled neural corpora for inferring neuronal identity from activity alone.
\vspace{2mm}
\end{itemize}

\begin{figure}[t]
    \centering
    \includegraphics[width=0.9\textwidth]{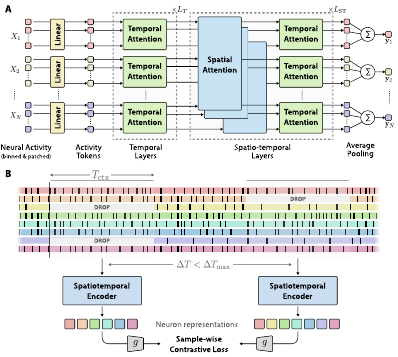}
    \caption{\footnotesize 
    \textbf{Overview of the NuCLR framework.} (A) The model takes as input the activity of a neural population over a fixed context window. Each neuron's activity is treated as a patched-token sequence and encoded across time using temporal transformer blocks. These temporally encoded tokens are then passed through spatio-temporal transformer layers that attends across neurons to incorporate population context. Finally, the model outputs one vector for each neuron in the population. (B) The resulting representations are trained with a sample-wise contrastive objective. 
    We sample two temporally-spaced views of the same population and encode them into neuron-level representations.
    Neurons are also randomly dropped before encoding to build robustness to partial observations. 
    Representations of the same neuron in both views are pulled closer, and all other neurons in the population are considered negatives and pushed apart.
    As a result, each neuron's representation is encouraged to be stable across time and distinguishable from other neurons in the population.
    \vspace{-4mm}
    }
    \label{fig:nuclr_architecture}
\end{figure}

%% file: files/methods.tex
\section{Method}
\label{sec:methods}

The goal of this work is to design and train deep learning models that, given the activity of a population of neurons, produce a latent representation for each neuron that captures its identity within the population.
Our approach is grounded in two core principles:
\begin{enumerate}[leftmargin=*, itemsep=1mm, topsep=1mm]
\item Capturing a neuron's identity requires not only its own activity but also the surrounding population activity, which provides essential contextual information.
\item A neuron's identity remains stable over time, 
so representations computed from different temporal segments of the population activity should be consistent with one another.
\end{enumerate}

Designing models that effectively incorporate population context is challenging because the number of recorded neurons varies widely across experiments, individuals, and sessions, making it non-trivial to define a consistent input structure or population-level operation.
Recordings also span multiple experimental conditions---including different stimuli and behaviors---which makes it hard for a model to generalize across recording sessions.
Together, these sources of variability make it difficult to learn neuron-level representations that are truly general-purpose and transferable to new recordings.
To satisfy our principles while accommodating these sources of variability, we introduce a model architecture (\cref{sec:architecture}) and training objective (\cref{sec:loss}).

\subsection{Model architecture}
\label{sec:architecture}

We aim to learn a function that maps the collective activity of a neural population to general-purpose neuron-level representations.
To support learning across recordings from different sessions and animals, each with a varying number of neurons, this function must be applicable to populations of arbitrary size.
Moreover, we do not assume any relational bias between neurons, and treat them as a permutation-equivariant set where all neuron-neuron interactions are modeled.
Finally, to enable representations that reflect a neuron's role within the circuit, the function should allow for information exchange among neurons, allowing each one to contextualize its activity relative to the population.
We formalize this as a set-to-set transformation:
\begin{equation}
\label{eq:gen-transformation}
\bigl\{\mathbf{y}_1, \mathbf{y}_2, \ldots, \mathbf{y}_N\bigr\} 
=
\mathcal{F}\Bigl(\bigl\{X_1, X_2, \ldots, X_N\bigr\}\Bigr),
\end{equation}
where $X_n$ denotes the spike-train sequence of the $n$-th neuron, and $\mathbf{y}_n$ is its corresponding representation.
We implement $\mathcal{F}$ using attention-based transformer blocks (\cref{fig:nuclr_architecture}A), which naturally support set-to-set mappings through their permutation-equivariant structure \cite{lee2019set}.

Given spike trains of a population of $N$ neurons over a temporal context window $T_\text{ctx}$, we begin by binning the spike train%
\footnote{
We use a fixed bin-size of 20ms, and also present a sweep over this value in \cref{sec:bin_size}. 
We also attempted a spike-tokenization based approach, and discuss it in \cref{sec:spike-tokenization}.
}, and partitioning the bins into non-overlapping temporal patches of length $T_\text{patch}$. 
The resulting binned-and-patched activity of $n$-th neuron is a sequence of vectors 
\mbox{$X_n = \bigl(\mathbf{x}_{n,1}, \mathbf{x}_{n,2}, \ldots, \mathbf{x}_{n,P} \bigr)$},
where $P = T_\text{ctx} / T_\text{patch}$ is the number of patches. 
Each patch is linearly projected into a $D$-dimensional latent space, yielding a token sequence 
\mbox{$Z^{(0)}_n = \bigl(\mathbf{z}^{(0)}_{n,1}, \ldots, \mathbf{z}^{(0)}_{n,P}\bigr)$}
for each neuron.

These latent tokens are first processed independently per neuron using 
$L_T$ layers of self-attention, which we refer to as \textbf{temporal transformer layers}. 
Each temporal layer $\mathcal{T}_\text{temp}$ operates on the patch sequence of a single neuron:
\begin{equation}
\label{eq:temporal-layer}
\bigl(\mathbf{z}^{(l+1)}_{n,1}, \ldots, \mathbf{z}^{(l+1)}_{n,P}\bigr) 
=
\mathcal{T}_\text{temp}
\left(
    \bigl(\mathbf{z}^{(l)}_{n,1}, \ldots, \mathbf{z}^{(l)}_{n,P}\bigr)
\right),
\quad \forall \; n \in [N].
\end{equation}
Within these layers, the temporal structure is maintained by the use of rotary position embeddings \cite{su2023roformerenhancedtransformerrotary,azabou2023unified}, which encode the relative timing of each patch without requiring absolute positional indices.

Following the stack of temporal layers, we apply $L_\text{ST}$ layers of \textbf{spatio-temporal transformer layers}, which alternate between spatial and temporal attention (\cref{fig:nuclr_architecture}A). 
In the spatial transformer layer, tokens at the same time index interact across the population via a shared transformer $\mathcal{T}_\text{spatial}$:
\begin{equation}
\label{eq:spatial-layer}
\bigl\{
    \mathbf{z}^{(l+1)}_{1,p}, \ldots, \mathbf{z}^{(l+1)}_{N, p}
\bigr\} =
\mathcal{T}_\text{spatial}
\left(
    \bigl\{
    \mathbf{z}^{(l)}_{1,p}, \ldots, \mathbf{z}^{(l)}_{N, p}
    \bigr\}
\right)
\quad \forall \; p \in [P].
\end{equation}
No positional embeddings were used in the spatial blocks. 
The temporal blocks in these layers reuse the structure defined in \cref{eq:temporal-layer}. So, $\mathcal{T}_\text{spatiotemporal} = \mathcal{T}_\text{spatial} \circ \mathcal{T}_\text{temp}$.

Finally, to obtain a fixed-dimensional representation for each neuron, we apply mean pooling over the temporal axis:
\begin{equation}
\mathbf{y}_n = \frac{1}{P} \sum_{p=1}^P \mathbf{z}^\text{final}_{n, p}
\quad \forall \; n \in [N].
\end{equation}

The core idea behind this architecture is to first build a temporally informed representation of each neuron based solely on its own activity. This is accomplished through the initial stack of temporal transformer layers, which process each neuron's token sequence independently. The resulting latent sequence captures the internal dynamics of each neuron over time\footnote{
The initial temporal-only layers, while not essential for performance, contribute to improved computational efficiency. Please see \cref{app:temporal-layers} for a longer discussion.}.
Next, the spatial transformer layers allow neurons to exchange information at each timepoint, injecting population-level context into each neuron's representation.
By alternating spatial and temporal layers, information received from other neurons at a given timepoint can be distributed to all other timepoints, creating more informed tokens to query the population again through the next spatial layer. 

This architecture can also be readily adapted for calcium imaging data, with minimal modifications (see \cref{app:adapt-to-calcium}). Additional details are provided in \cref{app:model-extra}.

\paragraph{What is considered a \textit{population}?}
For electrophysiology datasets, we treat each probe insertion as a distinct population, both for the spatial transformer layers and for the contrastive loss, where negative pairs are limited to neurons recorded on the same insertion (i.e. a single Neuropixels \cite{steinmetz2021neuropixels} probe).
In early experiments, we observed that allowing interactions across insertions led the model to cluster neurons based on probe identity rather than biologically meaningful properties.
This is undesirable, as our goal is to produce representations that reflect intrinsic neuronal attributes such as brain region and cell type.

\subsection{Self-supervised training objective}
\label{sec:loss}
Our goal is to learn the transformation $\mathcal{F}$ in \cref{eq:gen-transformation} such that it produces neuron-level representations that capture biologically meaningful properties such as cell type and brain region.
Since such labels are costly and difficult to obtain at scale, we adopt a self-supervised approach.
\texttt{NuCLR} leverages the natural spatial and temporal structure present in neural population activity and applies a contrastive learning objective to train $\mathcal{F}$ without requiring any explicit labels.

We begin by sampling two $T_\text{ctx}$-long windows, or \textit{views}, of neural population activity within  $\Delta T_{\max}$ of each other (\cref{fig:nuclr_architecture}B).  
Because identity-related properties such as cell type or brain region remain static, we encode this temporal-invariance using a contrastive loss that encourages corresponding neurons across the two views to produce similar representations.
At the same time, the representation should capture each neuron's distinct identity, so representations of different neurons within a population are pushed apart.

To promote robustness to partial observations and prevent overfitting to specific populations, we apply \emph{neuron dropout} independently to each view, randomly removing \emph{up to} 50\% neurons per view (\cref{sec:view-sampling}).
As a result, the number of neurons present in each view usually differs, and only a subset of neurons in one view have a corresponding neuron in the other view.
We define the indices of neurons presented in both views as
\begin{equation}
\mathcal{M} = \{(n, m) \mid \text{neuron } n \text{ in view 1 corresponds to neuron } m \text{ in view 2}\},
\end{equation}
which identifies all valid positive pairs for contrastive training.

Let $\tilde{\mathcal{X}}^1 = \{ X^1_1,\ldots,X^1_{N_1} \}$ and $\tilde{\mathcal{X}}^2 = \{ X^2_1,\ldots,X^2_{N_2} \}$ denote the two augmented views after neuron dropout.
The encoder maps these views into corresponding sets of representations:
\begin{align}
\mathcal{F}(\tilde{\mathcal{X}}^1) = \{ \mathbf{y}^1_1,\ldots,\mathbf{y}^1_{N_1}  \}
\;\;\text{and}\;\; 
\mathcal{F}(\tilde{\mathcal{X}}^2) = \{ \mathbf{y}^2_1,\ldots,\mathbf{y}^2_{N_2}  \}.
\end{align}
Similar to SimCLR~\cite{chen2020simple}, these representations are passed through a projection head $g(\cdot)$---a one-hidden-layer MLP---to obtain projected vectors $\mathbf{p}^v_n = g(\mathbf{y}^v_n)$, where $v \in \{1,2\}$ indexes the view.
We then apply an InfoNCE-based contrastive loss over the set of valid positive pairs $\mathcal{M}$:
\begin{align} 
\label{eq:loss-sample}
\mathcal{L}(\mathcal{F}, g \mid \tilde{\mathcal{X}}^1, \tilde{\mathcal{X}}^2) = 
& \frac{1}{|\mathcal{M}|}\sum_{(n,m) \in \mathcal{M}}
- \log \Big(
\frac
{\exp\left(\langle \mathbf{p}^1_n, \mathbf{p}^2_m \rangle / \tau \right)}
{
\displaystyle\sum_{n' \neq n}
\exp\left( \langle \mathbf{p}^1_n, \mathbf{p}^1_{n'} \rangle / \tau \right)
+
\sum_{(n, k) \notin \mathcal{M}}
\exp\left( \langle \mathbf{p}^1_n, \mathbf{p}^2_k \rangle / \tau \right)
}
\Big) \notag \\
& + \text{ symmetric term for view 2 to 1}
% + & \sum_{(n,m) \in \mathcal{M}}
% - \log \left(
% \frac
% {\exp\left(\langle \mathbf{p}^1_n, \mathbf{p}^2_m \rangle / \tau \right)}
% {
% \displaystyle\sum_{m' \neq m}
% \exp\left( \langle \mathbf{p}^2_m, \mathbf{p}^2_{m'} \rangle / \tau \right)
% +
% \displaystyle\sum_{(k,m) \notin \mathcal{M}}
% \exp\left( \langle \mathbf{p}^1_k, \mathbf{p}^2_m \rangle / \tau \right)
% }
% \right),
\end{align}

where $\langle\cdot,\cdot\rangle$ denote cosine-similarity, and $\tau$ is a temperature hyperparameter.
This loss encourages representations of the same neuron to be close across views while pushing apart those of different neurons within the same population.
As a result, the model learns neuron-level representations that are temporally stable and discriminative with respect to their functional roles.

A key distinction from standard SimCLR~\cite{chen2020simple} lies in how we handle the minibatch setting.
In \texttt{NuCLR}, the contrastive loss is computed \emph{independently} within each view pair. That is, when training on minibatches containing samples from different animals or sessions, we do not treat neurons across samples as negatives.
This design choice avoids an overabundance of \emph{easy negatives} in the denominator of \cref{eq:loss-sample}, which can degrade contrastive learning performance \cite{kalantidis2020hard}.
As a result, the number of negatives per sample is limited to the neurons within a single recording (typically in the hundreds).
This motivates our omission of positive-pair similarity terms from the denominator—following the decoupled contrastive loss (DCL)~\cite{10.1007/978-3-031-19809-0_38}---which has been shown to improve performance in regimes with few negatives.

In practice, for a minibatch containing $B$ independently sampled view pairs (potentially from different recordings), the overall loss is computed as a weighted average across samples:
\begin{equation}
\label{eq:loss-batch}
\mathcal{L}_\text{batch}(\mathcal{F}, g) =
\frac{1}{\sum_{b=1}^B N_b}
\sum_{b=1}^B N_b \cdot \mathcal{L}(\mathcal{F}, g | \tilde{\mathcal{X}}_b^1, \tilde{\mathcal{X}}_b^2),
\end{equation}
where $\mathcal{X}_b^{1,2}$ is the $b$-th view pair, and $N_b$ is the number of valid positive pairs (i.e., neurons present in both views after dropout). The inclusion of the $N_b$ terms helps to deal with the imbalance otherwise created between views with very different numbers of valid positive pairs.
We provide additional details about the model, hyperparameter choices, and other training details in \cref{app:model-extra}.

%% file: files/results.tex
\section{Results}
\label{sec:results}

\subsection{Experimental  setup}
To test the quality of the learned representations, we follow the standard linear-evaluation protocol used in self-supervised learning \cite{chen2020simple,grill2020bootstrap}. We first pre-train the model with our contrastive objective (\cref{sec:loss}), then freeze the encoder and compute a single representation per neuron by sequentially sampling windows from each session and averaging the resulting outputs.
Using these frozen neuron representations, we train linear classifiers to evaluate performance on two downstream tasks: \textit{cell type} decoding and \textit{brain region} decoding.

\vspace{-3mm}
\paragraph{Evaluation strategies.}
For each downstream task, we perform evaluation across three generalization settings:
(1)~\textit{Transductive}, 
where the testing populations 
are seen during self-supervised pretraining, and partial labels from these populations are used to train the classification head;
(2)~\textit{Transductive zero-shot}, where the test populations are present during pretraining, but no labels from them are used when training the classifier;
and (3)~\textit{Inductive zero-shot}, 
where the test populations are entirely unseen during pretraining, and no fine-tuning is performed on the encoder or classifier before evaluation.
For all three settings, we use a linear head on the output representations for the classification probe. 

Settings (1) and (2) emulate \textit{offline} (or post-hoc) analysis of neural recordings, where pretraining is allowed on the test populations.
Setting (1) reflects a scenario where a subset of neurons in a recording is labeled, and the goal is to infer labels for the rest.
Setting (2) captures the case where the test population is collected but is not labeled, so it used for self-supervised pretraining but cannot be used to train the classification probe.
Setting (3) corresponds to the \textit{online} use-case, where a pretrained model must operate on completely new neural populations without any retuning. 
This final setting is the most stringent and directly tests the model’s ability to generalize in a truly out-of-the-box fashion.

\vspace{-3mm}
\paragraph{Baseline methods.} We compare \nuclr~to several baselines: 
\textbf{NeuPRINT} \cite{neuprint}, a population-context model that uses summary statistics of neighboring neurons and the recorded behavior; 
\textbf{NEMO} \cite{yu2025in}, a CLIP-style contrastive approach that encodes individual neurons using their activity autocorrelograms and waveform templates;
and \textbf{LOLCAT} \cite{schneider2023transcriptomic}, a supervised model that uses the interspike interval distribution for individual neurons and leverages trial structure in the recording.
We refer the reader to \cref{sec:baseline_implementation} for further details on our implementation of these methods.
Our main metric is the \textbf{macro F1-score}, which is robust to class imbalance and enables fair comparison across datasets with varying label distributions
\footnote{We attempted to compare with a recent method NeurPIR \cite{wu2025neuron} but were not able to receive code from the authors to reproduce their method.
}.

We note that not all methods can be tested across all settings: NeuPRINT learns a neuron embedding table through back-propagation so it cannot be tested inductively without re-training. NEMO is designed only for electrophysiology data and cannot be applied to Bugeon et al. which consists of calcium imaging recordings. Finally LOLCAT is a supervised method so the transductive zero-shot setting is not possible. The performance of baseline methods in such cases will be reported as ``N/A''.

\subsection{Cell type decoding}
We begin by evaluating the quality of \nuclr's representations on their ability to predict neuronal cell types.
The first dataset used to test our approach is the Allen Brain Observatory Visual Coding (VC) Neuropixels dataset \cite{siegle2021survey}, which contains 58 sessions, each from a different mouse. 
Each animal is presented visual stimuli ranging from  drifting gratings to natural images and video.
Within the 58 sessions, 16 include optotagged inhibitory neurons labeled as one of 3 subclasses: Pvalb, Sst, or Vip.
Since all labeled neurons within a session belong to the same subclass, we evaluate performance only in the two zero-shot settings
\footnote{\label{ft:transductive}The transductive setting cannot allow for proper testing in this case, as a ``perfect'' classifier only has to infer which of the previously seen sessions does the neuron belong to and the cell-type corresponding to that session.}.
In the \textit{transductive zero-shot} setting, all sessions are used during pretraining (including labeled ones), but the decoder is trained and tested on neurons from non-overlapping subsets of mice. In the more stringent \textit{inductive zero-shot} setting, pretraining is restricted to unlabeled sessions only.
Further details on our data splits and validation methodology are provided in \cref{app:data}.

\input{tables/cell_type}

The second dataset that we evaluate on is a spatial transcriptomics dataset from  Bugeon et. al. \cite{Bugeon2022}, which consists of calcium imaging recordings from 17 sessions across 4 mice, with cell types labeled as excitatory (E) or inhibitory (I), and inhibitory neurons further divided into 5 subclasses (Lamp5, Pvalb, Vip, Sncg, Sst). We test both binary classification (E vs. I) and five-way subclass classification. Additionally, in \cref{sec:bugeon_extra}, we also test an 11-way subclass classification.

As shown in Table~\ref{tab:celltype}, \texttt{NuCLR} achieves strong performance across both datasets and consistently outperforms all baselines in zero-shot settings. On the Allen VC dataset, \texttt{NuCLR} achieves a macro F1-score of 0.7218 in the transductive zero-shot setting and 0.7200 in the inductive setting—more than 0.29 F1 higher than the next best method (NEMO) in the inductive case. On the Bugeon dataset, \texttt{NuCLR} achieves a macro F1 of 0.6738 on inductive zero-shot E vs. I classification and 0.3938 on the more challenging five-way subclass task on a held-out subject. 
We also compared with POYO+ \cite{azabou2025multisession} on the Allen VC in (\cref{app:compare_poyo+}), a multi-session multi-task decoding method that learns a neuron-level embedding table, and found that \nuclr~outperforms these embeddings by a strong margin on cell type classification.

These results demonstrate that \texttt{NuCLR} produces stable, transferable neuron representations that generalize to previously unseen populations without retraining, enabling accurate zero-shot decoding of cell type identity across diverse datasets and experimental conditions.

\subsection{Brain region decoding}
For brain region identification, we evaluate on two electrophysiological datasets with well-curated anatomical annotations: the International Brain Laboratory (IBL) Brain-wide Map~\cite{international2023brain} and the Steinmetz et al. 2019 dataset~\cite{Steinmetz2019}.
The IBL dataset comprises recordings from 139 mice across approximately 700 probe insertions. Following the evaluation setup used by NEMO, we perform classification across 10 brain regions (Figure~\ref{fig:scaling}A).
The Steinmetz et al. dataset includes 39 recordings from 10 mice, with classification over four regions: HPF, MB, TH, and VIS.
Zero-shot test populations correspond to entirely unseen experimental sessions (and thus unseen probe insertions) in the IBL dataset, and to unseen subjects in the Steinmetz et al. dataset.
Further details on our evaluation methodology and data folds are provided in \cref{app:data}.

As shown in Table~\ref{tab:region}, \texttt{NuCLR} achieves the highest macro F1-scores across all evaluation settings.
In the inductive zero-shot regime on IBL, \texttt{NuCLR} reaches an F1-score of 0.53 compared to the next best performing method NEMO with a 0.38.
We also observe in Figure~\ref{fig:scaling}A that the representations produced by \texttt{NuCLR} are organized according to brain region in the IBL dataset.
Embedding visualizations for the remaining datasets are provided in \cref{sec:visualizations}.
\input{tables/brain_region}

\subsection{Data scaling and label efficiency}

\begin{figure}[h]
  \centering
  \includegraphics[width=0.98\textwidth]{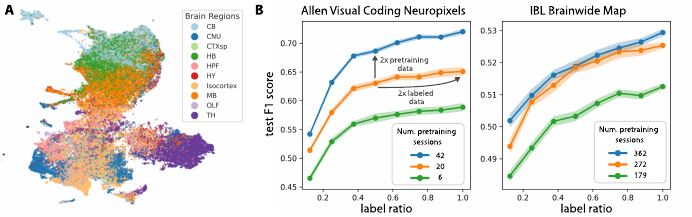}
  \caption{
  \footnotesize
  \textbf{Representation visualization and scaling trends.}
  (A) A 2-D UMAP visualization of \nuclr's neuron representations for the IBL Brainwide Map dataset colored by brain region.
  (B) Data scaling trends of inductive zero-shot decoding performance for cell type (Allen VC)  and brain region (IBL) tasks. The number of animals used in pretraining are varied along with the amount of labels used to train the linear classifier heads (label ratio).
  Performance improves with larger pretraining data and amount of supervision.
  In many cases, increasing the amount of unlabeled pretraining data is more effective than increasing the amount of labeled data.
  }
  \vspace{-4mm}
  \label{fig:scaling}
\end{figure}

We study how zero-shot performance scales with the amount of unlabeled pretraining data and the availability of labeled neurons for training the classification probe (\cref{fig:scaling}B).
We increase the amount of data used for pretraining while holding the test populations fixed and distinct from those used in pretraining.
To vary the level of supervision, we randomly subsample the labeled neurons used to train the linear classifier (reported as the ``label ratio''), where a ratio of 1.0 corresponds to using the same full set of labels across all pretraining scales.

We find that for the Allen VC dataset, increasing the number of unlabeled pretraining data leads to dramatic improvements in cell type classification performance.
In fact, in some cases, doubling the amount of pretraining data is significantly more beneficial than doubling the number of labeled neurons, as annotated in \cref{fig:scaling}B.
For the IBL dataset, we observe similar trends: brain region decoding performance continues to improve with additional pretraining data.

As expected, classification accuracy increases when more labeled data is used to train the classifier heads at a fixed pretraining data scale.
However, \nuclr~is highly label-efficient: when only 12.5\% of the labeled data is used, it still achieves scores 0.54 and 0.50 on Allen VC and IBL respectively, outperforming all baselines even when they use 100\% of the labels.
For instance, NEMO reaches only 0.42 and 0.38 under full supervision on these datasets.
Taken together, these results show that our approach is both label-efficient and highly scalable, capable of leveraging large unlabeled datasets to improve downstream performance with minimal supervision.

\subsection{Ablations}
\label{sec:ablations}
We conducted a series of ablation experiments to assess the contribution of key architectural and training components in our model (\cref{tab:ablations}). First, we removed the spatial attention layers which enable population-level interactions across neurons.\footnote{When ablating the spatial attention layers, the spatial layers were replaced with additional temporal layers to match the overall model depth and parameter count.} Ablating the spatial layers (population context) led to large performance drops on both the Allen VC (from 0.72 to 0.55) and IBL datasets (from 0.53 to 0.36). 
These results highlight the importance of spatial context: integrating information from surrounding neurons provides complementary signals to single-neuron activity and improves the model’s ability to capture features of neuronal identity.
It's interesting to note that even with this huge drop in accuracy, our model still outperforms most baselines in this restricted setting.

\begin{wraptable}{r}{0.65\textwidth}
\vspace{-4mm}
\centering
\caption{\footnotesize {\bf Ablation study on the Allen Visual Coding and IBL datasets.} 
We test the impact of spatial attention layers and neuron dropout on the Macro-F1 score. 
Values are reported as mean ± std.\ across 5 seeds.}
\label{tab:ablations}
\vspace{2mm}
\resizebox{0.65\textwidth}{!}{%
\begin{tabular}{lcc}
\toprule
\textbf{Model Variant} & \textbf{Allen VC (Cell type)} & \textbf{IBL (Brain region)} \\
\midrule
\textbf{Full NuCLR}                  & 0.7200\tpms{0.0267}  & 0.5295\tpms{0.0040}  \\
\quad w/o neuron dropout             & 0.6181\tpms{0.0252}  & 0.5248\tpms{0.0164} \\
\quad w/o spatial attn layers   & 0.5550\tpms{0.0796}  & 0.3573\tpms{0.0019}  \\
\bottomrule
\end{tabular}
}
\vspace{-3mm}
\end{wraptable}

We also examined the impact of neuron dropout, a regularization technique that randomly masks a subset of neurons during training.
This component improved performance on the Allen VC dataset but had minimal effect on IBL.
We attribute this difference to dataset scale: Allen VC pretraining dataset includes only 42 populations and therefore benefits more from regularization, whereas IBL comprises over 600 populations and may be less prone to overfitting.

%% file: tables/cell_type.tex
\begin{table}[t]
\centering
\caption{\textbf{Macro F1-score for cell type classification across different generalization settings.} Reported as mean ± std. dev. across 5 training seeds.
N/A indicates the method cannot operate in that evaluation setting.
\vspace{1mm}
}
\label{tab:celltype}
\resizebox{\textwidth}{!}{
\begin{tabular}{lc|c|cccc}
\toprule
\textbf{Dataset} & \textbf{\# Classes} & \textbf{Setting} & \textbf{NuCLR (Ours)} & \textbf{NeuPRINT} & \textbf{NEMO} & \textbf{LOLCAT} \\
\midrule
\multirow{2}{*}{Allen VC}\footref{ft:transductive} & \multirow{2}{*}{3} 
    & Transd. zero-shot & 0.7218\tpms{0.0113} & 0.4020\tpms{0.0238} & 0.4256\tpms{0.0114} & N/A \\
    &                 
    & Ind. zero-shot    & 0.7200\tpms{0.0267} & N/A & 0.4194\tpms{0.0099} & 0.4121\tpms{0.0800} \\
\midrule
\multirow{3}{*}{Bugeon (E vs. I)} & \multirow{3}{*}{2} 
    & Transductive
    & 0.8110\tpms{0.0035}
    & 0.6658\tpms{0.0090} 
    & N/A 
    & 0.7205\tpms{0.0127} \\
    &                 
    & Transd. zero-shot 
    & 0.6826\tpms{0.0293} 
    & 0.6362\tpms{0.0073}
    & N/A
    & N/A \\
    &                 
    & Ind. zero-shot
    & 0.6738\tpms{0.0563}
    & N/A
    & N/A
    & 0.7463\tpms{0.0095} \\
\midrule
\multirow{3}{*}{Bugeon (Subclass)} & \multirow{3}{*}{5} 
    & Transductive
    & 0.6101\tpms{0.0249} 
    & 0.4952\tpms{0.0222} 
    & N/A 
    & 0.2900\tpms{0.0388} \\
    &                 
    & Transd. zero-shot 
    & 0.4014\tpms{0.0268} 
    & 0.3529\tpms{0.0194} 
    & N/A 
    & N/A \\
    &                 
    & Ind. zero-shot    
    & 0.3938\tpms{0.0556}
    & N/A
    & N/A
    & 0.2418\tpms{0.0078} \\
\bottomrule
\end{tabular}
}
\vspace{-4mm}
\end{table}

%% file: tables/brain_region.tex
\begin{table}[t]
\centering
\caption{\footnotesize {\bf Macro F1-scores for brain region classification across different generalization settings.} 
Reported as mean ± std. dev. across 5 training seeds.
N/A indicates the method cannot operate in that evaluation setting.
\vspace{1mm}
}
\label{tab:region}
\resizebox{\textwidth}{!}{%
\begin{tabular}{lc|c|cccc}
\toprule
\textbf{Dataset} & \textbf{Classes} & \textbf{Setting} & \textbf{NuCLR (Ours)} & \textbf{NeuPRINT} & \textbf{NEMO} & \textbf{LOLCAT} \\
\midrule
\multirow{3}{*}{IBL} & \multirow{3}{*}{10} 
    & Transductive   & 0.6686\tpms{0.0034} & 0.2734\tpms{0.0153} & 0.4188\tpms{0.0041} & 0.2851\tpms{0.0008}  \\
    &                                                            
    & Transd. zero-shot & 0.5343\tpms{0.0115} & 0.2531\tpms{0.0137} & 0.3804\tpms{0.0011} & N/A \\
    &                 
    & Ind. zero-shot    & 0.5295\tpms{0.0040} & N/A & 0.3793\tpms{0.0011} &  0.2532\tpms{0.0016} \\
\midrule
\multirow{3}{*}{Steinmetz et. al.} & \multirow{3}{*}{4} 
    & Transductive           & 0.9594\tpms{0.0027} & 0.4476\tpms{0.0166} & 0.6989\tpms{0.0044} & 0.3205\tpms{0.0055} \\
    &                 
    & Transd. zero-shot & 0.7338\tpms{0.0226} & 0.4122\tpms{0.0326} & 0.6681\tpms{0.0016} & N/A \\
    &                 
    & Ind. zero-shot    & 0.5810\tpms{0.0110} & N/A & 0.5595\tpms{0.0048} & 0.3191\tpms{0.0311}  \\
\bottomrule
\end{tabular}
}
\vspace{-4mm}
\end{table}

%% file: files/relatedworks.tex
\section{Related Work}

\paragraph{Functional organization of brain circuits.}
From early days of neuroscience and experiments of Hubel and Wiesel \cite{hubel1959receptive}, mapping the responses or ``tuning'' of different neurons has been an important part of building our understanding of how the brain works. At scale, and with modern large-scale datasets, this has allowed for extensive work in characterizing the functional organization in primate V4 \cite{Willeke2023}, in chromatic feature detectors in retina \cite{Hfling2024}, visual cortex in mouse \cite{Tong2023}, and combinatorial codes in mouse V1 \cite{Ustyuzhaninov2022}. In vision, maximally exciting or most discriminative inputs \cite{burg2024most, Walker2019inception,Tong2023} can be used to find stimuli that help to differentiate neurons with different functional properties.  

\vspace{-3mm}
\paragraph{Deep learning approaches for cell type and brain region classification.}
Attempts to build deep learning solutions for the problem of extracting cell types and brain regions from neural activity are still nascent \cite{schneider2025robust}. 
In \cite{schneider2023transcriptomic,tolossa2024conserved}, interspike interval (ISI) distributions are used for cell type and brain region prediction, respectively. LOLCAT \cite{schneider2023transcriptomic} extracts the ISI distribution over individual trials and builds an multi-head attention layer to attend to subsets of trials and classify neurons in a supervised manner. 
NEMO \cite{yu2025in}, PhysMAP \cite{lee2024physmap} and VAE-based models \cite{beau2025deep} combine waveform and autocorrelogram views from a single neuron; NEMO uses a CLIP-style contrastive loss between both modalities, and Beau et al.\ concatenating both modalities after separate VAE-based encoders are applied to each.

Recent methods have begun to explore integration of population context into neuron embeddings. 
NeuPRINT \cite{mi2023learning} uses a  reconstruction task and stimulus information to learn an embedding table for neurons, and adds population context through a set of fixed summary statistics. 
NeurPIR \cite{wu2025neuron} learns time-invariant neuron representations from population dynamics using a contrastive VICReg loss and a pretrained population-level encoder to provide population context for each neuron. 
While these approaches offer improvements over single-neuron models, their reliance on compressed population features can discard fine-grained structure in the dynamics.
Additionally, because these models must be retrained or finetuned for each new recording, they provide limited ability to generalize in a zero-shot manner across sessions or animals.

\vspace{-3mm}
\paragraph{Channel-level transformer architectures and functional embeddings.}
A growing line of work applies transformer architectures at the channel or neuron level to generate embeddings from population activity \cite{azabou2023unified, azabou2025multisession, zhang2025neural, liu2022seeing, le2022stndt}. POYO and POYO+\footnote{While POYO+ reports Cre-line and brain-region classification, these results are obtained using session-level averaged latent representations, and are not based on embeddings of individual neurons. See Appendix~\ref{app:compare_poyo+} for more details and an extended analysis of the unit embeddings learned by POYO+.} 
learn a unit embedding table and EIT generates a channel-level tokenization; all focus on supervised behavior decoding tasks. NEDS uses both encoding (neural activity prediction) and decoding (behavior prediction) objectives, and learns a set of neuron level weights at the decoder which can be used to read out neuron-level attributes. STNDT employs a masked modeling objective with both neuron-level and population-level tokens to achieve strong performance on prediction of neural activity on held-out neurons. However, none of these approaches are designed to learn identity-relevant neuron embeddings or to generalize in zero-shot settings. \texttt{NuCLR} is specifically built for this purpose, with an objective and architecture tailored to recover neuron identity from population activity and transfer robustly across sessions and animals.

\vspace{-3mm}
\paragraph{Contrastive methods and their use in neural population data analysis.}
Contrastive learning maps similar examples to nearby points in an embedding space while separating dissimilar ones \citep{le2020contrastive, chen2020simple}, and has achieved broad success across language \citep{gao2021simcse}, vision \citep{chen2020simple}, audio \citep{saeed2021contrastive}, and multimodal representation learning \citep{radford2021learning}. Motivated by these advances, contrastive objectives have increasingly been used to learn representations from neural population activity \citep{azabou2021mine, liu2021drop, peterson2022learning, urzay2023detecting, schneider2023learnable}. Swap-VAE and MYOW introduce temporal and dropout-based augmentations for population responses \citep{liu2021drop, azabou2021mine}; Swap-VAE combines contrastive and generative losses across two latent spaces to learn disentangled population embeddings, while MYOW employs a BYOL-based \citep{grill2020bootstrap} contrastive objective with nearest-neighbor mining to identify harder positives. Urzay et al.\ apply contrastive metric learning \citep{ahad2022learning} to detect change points in neural time series \citep{urzay2023detecting}. Peterson et al.\ demonstrate cross-stream self-supervised learning between neural activity and behavior \citep{peterson2022learning}. CEBRA uses contrastive learning to align recordings across sessions, modalities, and animals  \citep{schneider2023learnable}. 
While these approaches successfully learn representations of neural activity that can be connected to behavior, they operate at the population level and do not construct embeddings for individual neurons. \texttt{NuCLR} applies contrastive learning at the neuron level: different temporal views of the same neuron serve as positives, all other neurons serve as negatives. This design enables \texttt{NuCLR} to learn identity-relevant embeddings for individual neurons in the context of other neurons, rather than global representations which compress the entire population into a single embedding.

%% file: files/discussion.tex
\section{Discussion}
\label{sec:discussion}
We introduced \texttt{NuCLR}, a self-supervised framework for learning population-aware {\em neuron identity embeddings} from both electrophysiology and calcium imaging datasets. Using a spatiotemporal transformer trained with a contrastive objective, \texttt{NuCLR} captures within-neuron dynamics and population context, and enables \emph{zero-shot} prediction of neuronal properties in new subjects without supervised labels or session-specific tuning.

Even though we don't use stimulus information in our pretraining, our current evaluations use datasets with similar underlying task structure. Thus, testing and building robustness to changes in the underlying stimulus distribution will be an interesting line of future research.  As we extend the approach to train on datasets spanning different tasks, sensory modalities, and behavioral states, it may be useful to incorporate task-conditioned encoders, or metadata-aware prompts to support training across heterogeneous datasets.

While our current evaluations show the applicability of \texttt{NuCLR} on both electrophysiology and calcium imaging, we currently need two different models for each modality. In the future, building a unified pretraining approach for both modalities \cite{knight2025unified} would allow for a joint model pretraining that could take advantage of both modalities and train at an even larger scale than with a single modality alone.

Our zero-shot scaling results show strong benefits of pretraining with more data on more animals. Thus, we anticipate that \texttt{NuCLR} could be trained on even more data to achieve even more performance gains, and ultimately building a path toward general-purpose neuron identity estimation and a foundation for models that integrate across modalities, species, and experimental settings. 
As larger and more diverse neural datasets emerge, such approaches may help reveal how neuronal diversity shapes computation across brain-wide circuits.

%% file: files/funding.tex
\section*{Acknowledgments and Disclosure of Funding}
We would like to thank Anna Lakunina, Shivashriganesh P. Mahato, Alexandre André, and Liam Paninski for insightful discussions and feedback.
This project was supported by the NIH (1R01EB029852), the NSF   (CAREER IIS-2146072),  as well as generous gifts from the CIFAR Learning in Machines \& Brains Program and The Hypothesis Fund.

%% file: appendix/main.tex
\appendix
\section*{Appendix}

\input{appendix/more_results}
\input{appendix/model}

\input{appendix/visualizations}
\input{appendix/datasets}
\input{appendix/baselines}

%% file: appendix/more_results.tex
\section{Additional Results}
\label{app:extra-results}

\subsection{Effect of bin-size}
\label{sec:bin_size}
The encoder used in this work assumes the neural activity is binned with a finite bin-size, which sets the minimum time resolution at which the encoder can views the data itself. A fair hypothesis would be that certain cell-types or neurons from certain brain-regions would be more (or less) identifiable at a certain bin-size setting.

\begin{figure}[h]
  \centering
  \includegraphics[width=0.95\linewidth]{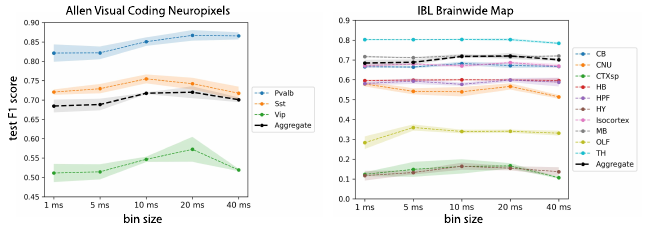}
  \caption{\textbf{Results of bin-size sweep.} Classification performance variation for different classes upon sweeping the bin-size hyperparameter of the encoder. ``Aggregate'' refers to the overall macro-F1 score. Error polygons represents standard error of mean (SEM) measured across 3 pretraining seeds.}
  \label{fig:bin-size-sweep}
\end{figure}

We test this by performing a sweep over the bin-size, and measuring the class-wise and macro F1-scores.
As we see in \cref{fig:bin-size-sweep}, some kinds of neurons do indeed prefer certain bin-sizes, however, the overall effect of this hyperparameter is surprisingly small. In other words, the overall classifier performance is relatively robust to the choice of bin-size.

\subsection{Experiments with a spike-based encoder}
\label{sec:spike-tokenization}
\begin{figure}[h]
  \centering
  \includegraphics[width=\linewidth]{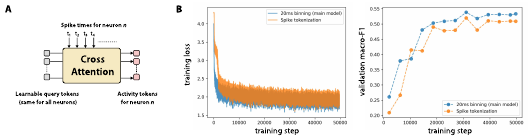}
  \caption{
  \textbf{Spike-tokenization based encoder.}
  (A) A spike-tokenization layer for the \nuclr\ encoder. 
  (B) Pretraining-time metrics comparison for spike-tokenization version and binning version of \nuclr. These plots are for pretraining on the IBL dataset.}
  \label{fig:spike-tokenization}
\end{figure}

We also explored a spike-tokenization strategy inspired by POYO \cite{azabou2023unified}.
Unlike POYO, our model aims to \textit{produce} neuron-level embeddings at its outputs, and therefore cannot assume or learn neuron identities through separate “unit embeddings.” Instead, we assign the same learnable embedding to all neurons, so that each spike token consists only of its timestamp and this shared embedding. 
Each neuron's spike train is processed using perceiver-style cross-attention layers, which are queried by learnable vectors placed at linearly spaced timestamps---using a spacing equal to the patch length in our main model. The shared embedding serves only to enable cross-attention over the spike sequence and does not encode any neuron-specific information.
The resulting set of tokens is then treated as activity tokens (as shown in \cref{fig:nuclr_architecture}A) and passed through the spatio-temporal transformer identically to our main architecture.
A schematic of this spike-tokenization design is shown in \cref{fig:spike-tokenization}A.

We find that this spike-based encoder performs significantly worse on the Allen VC cell-type classification task, while achieving nearly comparable performance on the IBL brain-region classification task (\cref{tab:ablation-spikes}).
We also note that the training stability and convergence speed is much better in the binned-and-patched version of the model, as shown in \cref{fig:spike-tokenization}B.
Alternative implementations of spike tokenization may yield improved results, however, we leave that for future work.

\begin{table}[h]
\centering
\caption{\footnotesize {\bf Spike-tokenization ablation.} Values are reported as mean ± std. dev. across 5 seeds for the 20ms binning model (main \nuclr), and 3 seeds for the spike tokenization based model.
\vspace{1mm}
}
\label{tab:ablation-spikes}
\resizebox{0.6\textwidth}{!}{%
\begin{tabular}{lcc}
\toprule
\textbf{Model Variant} & \textbf{Allen VC} & \textbf{IBL} \\
\midrule
20ms binning (main \nuclr)
& 0.7200\tpms{0.0267}  
& 0.5295\tpms{0.0040}  \\
Spike-tokenization based	
& 0.6665\tpms{0.0167}
& 0.5268\tpms{0.0136} \\
\bottomrule
\end{tabular}
}
\end{table}

\subsection{Probing intermediate layer representations}
\begin{figure}[h]
  \centering
  \includegraphics[width=0.9\textwidth]{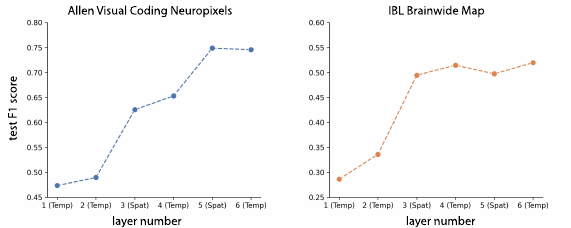}
  \caption{\footnotesize\textbf{Classification performance of intermediate layer representations.} Plots present the classification performance at the \textit{output} of all layers in our 6 layer model for a single seed.}
  \label{fig:intermediate-layers}
\end{figure}

In all our experiments, we used a 6-layer implementation of \nuclr, with the first two layers being temporal-only layers followed by two spatio-temporal layers (each having 2 transformer blocks). 
In \cref{fig:intermediate-layers} we measure the cell-type and brain-region decoding accuracies of the embeddings obtained at all 6 layers.
The results show a general improvement of classification performance as we go deeper in the network, with major performance jumps being observed whenever the representation passes through spatial layers. This is another confirmation of the importance of spatial layers, as shown in \cref{sec:ablations}.

\subsection{Comparison with POYO+ on Allen VC}
\label{app:compare_poyo+}
POYO+ \cite{azabou2025multisession} is a multi-task behavior decoder model that can be trained simultaneously on multiple recordings. It has been shown to perform well on a Cre-line classification task on the Allen Brain Observatory calcium imaging dataset, using the \textit{session-averaged} latent outputs of its encoder. This is possible since each recording (session) in that dataset has neurons with only one cell-type. This analysis method used in the POYO+ manuscript is not directly applicable here since we aim to produce neuron-level embeddings that can identify different types of neurons \textit{within} a given population.

However, POYO+ learns a ``unit embedding'' for each neuron, which can, in theory, be used for the purpose of decoding neuron-level properties. We use this strategy to compare POYO+ with \nuclr.
We train POYO+ on the Allen VC dataset for decoding the following behaviors: Drifting Gratings Orientation, Drifting Gratings Temporal Frequency, Gabor Orientation, Gabor Position, Natural Scenes, Running Speed, and Static Gratings Orientation. 

Since POYO+ can only be tested under transductive evaluations, we choose the transductive zero-shot regime for our comparison, and find that \nuclr~outperforms the learned unit embeddings of POYO+ by a strong margin (\cref{tab:poyo+}).
\begin{table}[h]
\vspace{-2mm}
\centering
\caption{\textbf{Comparing the unit embeddings in POYO+ with \nuclr~and other methods.}
Evaluation done in transductive zero-shot setting on the Allen VC dataset, with result being macro-F1 scores represented as mean $\pm$ std. dev. over 5 seeds. 
\vspace{1mm}
}
\label{tab:poyo+}
\resizebox{0.55\textwidth}{!}{%
\begin{tabular}{cccc}
\toprule
 \textbf{POYO+} & \textbf{NuCLR} & \textbf{NeuPRINT} & \textbf{NEMO} \\
\midrule
 0.3521\tpms{0.0233} & $0.7218\tpms{0.0113}$ & $0.3999\tpms{0.0312}$ & $0.4256\tpms{0.0114}$ \\
\bottomrule
\end{tabular}
}
\end{table}

Pretraining POYO+ required an average of 7.5 hours on 4 NVidia B200s to converge (about 100 epochs), while \nuclr~requires only about 1.5 hours on the same hardware. We used the main training configuration from the example provided by the authors\footnote{\href{https://github.com/neuro-galaxy/torch_brain/tree/main/examples/poyo_plus}{\texttt{https://github.com/neuro-galaxy/torch\_brain/tree/main/examples/poyo\_plus}}}.
We also report the behavior decoding performance achieved by POYO+ after training across five random seeds:
\begin{itemize}[leftmargin=5mm]
\item {Drifting Gratings Orientation Accuracy:} 93.07\%\tpms{1.76\%} (Chance: 12.5\%)
\item {Drifting Gratings Temporal Frequency Accuracy:} 93.40\%\tpms{1.36\%} (Chance: 20\%)
\item {Gabor Orientation Accuracy:} 56.12\%\tpms{1.18\%} (Chance: 25.0\%)
\item {Gabor Position (2D) $R^2$:} 0.6888\tpms{0.0487}
\item {Natural Scenes Accuracy:} 53.23\%\tpms{10.18\%} (Chance: 0.84\%)
\item {Running Speed $R^2$:} 0.7681\tpms{0.0115}
\item {Static Gratings Orientation Accuracy:} 75.99\%\tpms{0.97\%} (Chance: 12.5\%)
\end{itemize}

\subsection{Ablating temporal-only attention layers}
\label{app:temporal-layers}
To assess the importance of dedicated temporal-only attention layers, we replace them with an equivalent number of spatio-temporal layers. Specifically, the first two temporal-only layers in the original model were substituted with one spatio-temporal layer to maintain a similar model capacity.

As shown in \cref{tab:ablation-st} the inclusion of dedicated temporal-only layers appears to be a relatively inconsequential design decision, as performance does not change significantly between the two model configurations.
However, a benefit of retaining the temporal-only layers is computational efficiency.
Given that the number of temporal patches ($\sim10$) is typically much smaller than the number of neurons ($\sim100$), a temporal layer has less computationally expensive than a 
using separate temporal layers reduces the overall computational complexity, as attention mechanisms scale quadratically with the number of tokens in a sequence.

\begin{table}[h]
\centering
\caption{\footnotesize {\bf Ablation study for temporal-only layers.} Values are reported as mean ± std. dev. across 5 seeds for full \nuclr~, and 3 seeds for the ablated model.
\vspace{1mm}
}
\label{tab:ablation-st}
\resizebox{0.555\textwidth}{!}{%
\begin{tabular}{lcc}
\toprule
\textbf{Model Variant} & \textbf{Allen VC} & \textbf{IBL} \\
\midrule
\textbf{Full NuCLR}                  
& 0.7200\tpms{0.0267}  
& 0.5295\tpms{0.0040}  \\
\quad w/o temporal attention layers 
& 0.7184\tpms{0.0124}
& 0.5259\tpms{0.0089} \\
\bottomrule
\end{tabular}
}
\end{table}

\subsection{Confusion matrices}
\label{app:confusion_matrices}
\begin{figure}[h]
    \centering
    \includegraphics[width=0.9\linewidth]{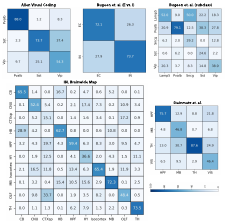}
    \caption{
    \textbf{Confusion matrices achieved by \nuclr~on inductive zero-shot evaluation.}}
    \label{fig:confusion-matrices}
\end{figure}

Please refer to \cref{app:confusion_matrices} for confusion matrices.
We present the confusion matrices achieved by \nuclr~in \cref{fig:confusion-matrices} for all datasets evaluated in the inductive zero-shot setting.

\subsection{Additional results on Bugeon et. al. dataset}
\label{sec:bugeon_extra}

We evaluate the representations of \nuclr~and baseline models on an 11-class label set of the Bugeon et. al. dataset. This label set consists of the classes: 
Lamp5-Chrna7, Lamp5-Npy, Lamp5-Tmem182, Pvalb-Tac1, Pvalb-Vipr2, Sncg-Pdzrn3, Sncg-Vip, Sst-Reln, Sst-Tac1, Vip-Cp, Vip-Reln.
As seen in \cref{tab:bugeon_extra}, \nuclr~outperforms both NeuPRINT and LOLCAT baselines on this labelling of neurons.

\input{tables/bugeon_extra}

%% file: tables/bugeon_extra.tex
\begin{table}[h]
\centering
\caption{\textbf{Macro F1-score for cell type classification on the 11-class label set of Bugeon et. al.} Reported as mean ± std. dev. across 5 training seeds.
N/A indicates the method cannot operate in that evaluation setting.
\vspace{1mm}
}
\label{tab:bugeon_extra}
\resizebox{\textwidth}{!}{
\begin{tabular}{lccccc}
\toprule
\textbf{Dataset} & \textbf{\# Classes} & \textbf{Setting} & \textbf{NuCLR} & \textbf{NeuPRINT} & \textbf{LOLCAT} \\
\midrule
\multirow{3}{*}{Bugeon et. al. (11-class)} & \multirow{3}{*}{11} 
    & Transductive          
    & 0.4056\tpms{0.0425}
    & 0.2748\tpms{0.0131}
    & 0.1559\tpms{0.0318}
    \\
    &                 
    & Transductive zero-shot 
    & 0.2333\tpms{0.0129}
    & 0.1825\tpms{0.0158}
    & N/A 
    \\
    &                 
    & Inductive zero-shot    
    & 0.1990\tpms{0.0591}
    & N/A
    & 0.1482\tpms{0.0256} \\
\bottomrule
\end{tabular}
}
\end{table}

%% file: appendix/model.tex
\section{Additional method details}
\label{app:model-extra}

\subsection{View-pair sampling}
\label{sec:view-sampling}
We sample view pairs during pretraining in three steps:
\begin{enumerate}[leftmargin=5mm]

\item{\textbf{First view.}} The first view is sampled using the \texttt{RandomFixedWindowSampler} found in \texttt{torch\_brain}\footnote{\href{https://github.com/neuro-galaxy/torch_brain}{\texttt{https://github.com/neuro-galaxy/torch\_brain}}}.
This sampler divides each recording into non-overlapping windows of length $T_\text{ctx}$ and randomly samples one window (without replacement throughout an epoch).
Across epochs, a shared random jitter to the window start times at the beginning of every epoch. This emulates uniform sampling while ensuring complete data coverage each epoch.

\item{\textbf{Second view.}} The second view is sampled relative to the first.
Its start time is drawn uniformly from a range constrained by the $\Delta T_\text{max}$ setting, ensuring the two views come from nearby temporal contexts.

\item{\textbf{Neuron dropout.}}
To increase robustness to partial observations, we apply neuron dropout independently to each view.
Given a view with $N$ neurons, we first sample the number of neurons to drop from a uniform distribution: $N_\text{drop} \sim \mathcal{U}(0, 0.5 N)$.
We then randomly select $N_\text{drop}$ neurons and exclude them from the view during that training step.
\end{enumerate}

\subsection{Model implementation details}

\paragraph{Transformer.}
We use standard scaled-dot-product attention as implemented in xformers \cite{xFormers2022},
pre-normalization using LayerNorm. Our feedforward network (FFN) uses the GEGLU activation function, with its hidden dimension being 4x the dimension of the tokens ($D$).
In the temporal transformer layers, we use Rotary Embeddings to incorporate timing information, as described below.

\paragraph{Rotary Time Embeddings.}
We use rotary time embeddings following the formulation in Section A.1 of~\cite{azabou2023unified}, which includes the use of value rotation in addition to query-key rotation.
The only difference in our implementation lies in the choice of temporal scaling parameters.
Because our input tokens are uniformly spaced with a stride of $T_\text{patch}$, we set $T_\text{min} = T_\text{patch}$ and $T_\text{max} = 8 \times T_\text{ctx}$.

\subsection{Adapting to Calcium Imaging Data}
\label{app:adapt-to-calcium}
To adapt \texttt{NuCLR}'s spatio-temporal transformer to calcium imaging data, we replace the binning-and-patching step used for spike trains with a temporal patching operation applied directly to the calcium fluorescence time series (i.e. the $\Delta F/F$ signal).
This modification affects only the input stage of the model; the architecture and training procedure remain unchanged.
Unlike electrophysiology, calcium imaging does not involve physical probes or insertions. Therefore, we treat all simultaneously recorded neurons within a session as a single population—both for the spatial transformer layers and for contrastive loss computation.

\subsection{Hyperparameters}
We use the AdamW \cite{loshchilov2018decoupled} optimizer with a linear learning rate warm-up over the first epoch, followed by cosine decay until end of training.
All relevant hyperparameters for training \texttt{NuCLR} are listed in \cref{tab:hyperparameter}.
These values were mainly selected via manual line searches on the IBL development set (for ephys data, \cref{sec:ibl_data_detail}) and the Bugeon et al.~development set (for calcium imaging data, \cref{sec:bugeon_data_detail}).
Across all datsets, we pretrain the model for 50{,}000 steps and use the \texttt{bfloat16} number format throughout. Pretraining takes approximately 3 hours on a machine with 4 $\times$ NVidia H100 GPUs.

\begin{table}[ht]
\vspace{-5mm}
\caption{
\textbf{Key hyperparameters for \nuclr.}
\vspace{1mm}
}
\label{tab:hyperparameter}
\centering
\resizebox{0.65\textwidth}{!}{%
\begin{tabular}{lll}
\toprule
\textbf{Parameter} & \textbf{Value for Ephys.}  & \textbf{Value for Ca$^{+2}$} (diff. only) \\
\midrule
$T_{\text{ctx}}$        & 10s                & 30s                   \\ 
$T_{\text{patch}}$      & 1s                 &                       \\ 
$\Delta T_{\text{max}}$ & 30s                & 240s                  \\ 
Bin size                & 20ms               & N/A                   \\ 
\midrule
$L_{\text{T}}$          & 2                  &                       \\ 
$L_{\text{ST}}$         & 2                  &                       \\ 
$D$                     & 256                &                       \\ 
Num. attention heads    & 4                  &                       \\
\midrule
Linear dropout          & 0.2                &                       \\ 
Attention dropout       & 0.0                &                       \\ 
Max. neuron dropout     & 50\%               &                       \\ 
\midrule
Num. training steps     & 50,000             &                       \\ 
Batch size              & 128                & 16                    \\ 
Max learning rate       & $5 \times 10^{-4}$ & $1.25 \times 10^{-4}$ \\ 
Weight decay            & 0.01 (default)     &                       \\
$\beta_1$               & 0.9 (default)      &                       \\
$\beta_2$               & 0.999 (default)    &                       \\
\bottomrule
\end{tabular}
}
\end{table}

%% file: appendix/visualizations.tex
\section{Embedding Visualizations}
\label{sec:visualizations}
\begin{figure}[t]
    \centering
    \includegraphics[width=\linewidth]{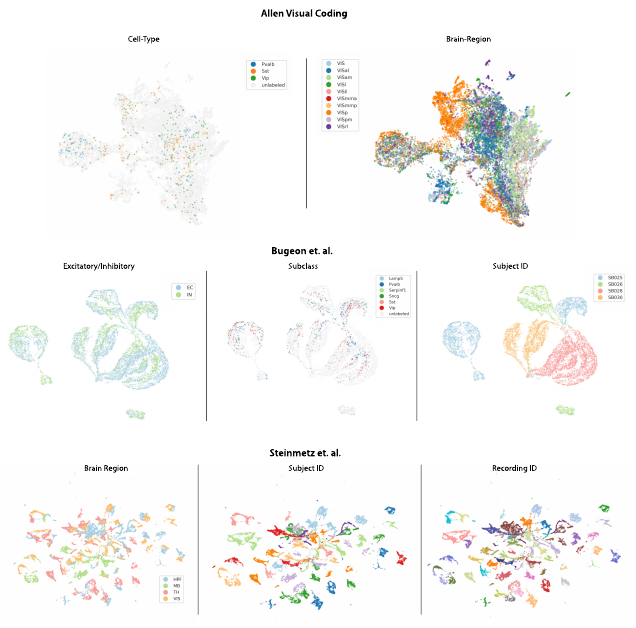}
    \caption{\textbf{UMAP visualizations} of \nuclr's embeddings colored by various properties.}
    \label{fig:visualizations}
\end{figure}

We present UMAP-based visualizations of \nuclr's output representations in \cref{fig:visualizations}.
For the Allen VC dataset, which includes a large number of subjects, the embeddings exhibit clear density modes that align most strongly with brain-region information.

In contrast, the Bugeon et al. and Steinmetz et al. datasets are relatively small, containing only 4 and 10 subjects respectively, with limited total number of recordings. For these datasets, \nuclr's embeddings cluster primarily by subject or session identity. 
However, \textit{within} these clusters, we still observe meaningful density modes corresponding to cell-type and brain-region structure.
Embeddings for the IBL dataset, shown in \cref{fig:scaling}A, reveal strong region-based organization without noticeable clustering by subject or session.

The subject- and session-specific clustering observed in smaller datasets (Bugeon et al. and Steinmetz et al.) may hinder data-driven discovery, as it suggests entanglement with recording-specific factors. While this effect appears only in small datasets, mitigating it remains an important direction for future work.

%% file: appendix/datasets.tex
\section{Details on datasets and evaluation methodology}
\label{app:data}

This section contains description of each dataset, and details our train-test splits for all evaluation settings.
Additionally, our code-base\footnote{\codelink} includes all preprocessing, split preparation, and evaluation scripts. 

\subsection{Allen Visual Coding}
\label{app:data-allenvc}

The Allen Visual Coding (VC) dataset consists of 58 Neuropixels recordings, each from a unique mouse.
Each recording spans approximately two hours, during which various visual stimuli are presented.
Of these recordings, 16 contain optotagged neurons,\footnote{We use the same set of labeled neurons as in the original LOLCAT publication~\cite{schneider2023transcriptomic}.}
with each labeled recording containing neurons from only one of the three inhibitory cell types: Pvalb, Vip, or Sst.
Since reliable cell-type labels are only available for neurons in the visual cortex, we restrict all models—including \texttt{NuCLR} and baselines—to use only neurons from the VIS region during pretraining.

For zero-shot classifier evaluation, we follow a leave-one-subject-out strategy. Specifically, there are 16 test folds corresponding to the 16 labeled subjects.
In each fold, one subject is held out for testing, and the remaining 15 are used for training.
Within the training set, we perform 4-fold cross-validation (subject-wise) to select the best training epoch.
The final test score is reported from a classifier trained on all 15 subjects using the selected epoch.
We did not test classifier performance on this dataset in the transductive setting as a ``perfect'' classifier only has to infer which of the previously seen sessions does the neuron belong to, and the cell-type corresponding to that session.

\subsection{IBL Brainwide Map}
\label{sec:ibl_data_detail}

The IBL Brainwide Map dataset consists of 439 recordings from 139 unique mice, with each recording performed using one or two Neuropixels insertions.
While performing zero-shot evaluations (both inductive and transductive), we construct train–test splits at the recording level, ensuring that no co-recorded neuronal populations appear in both sets.
In total, 93 recordings are designated for the test set, and the remaining 346 are used for training and validation.
Of the training set, 91 recordings are used as a development set for tuning hyperparameters specific to electrophysiology data.
During classification, these same 91 recordings also serve as a validation set for selecting the best training epoch.
Final performance metrics are reported using a model trained on all 346 training recordings, evaluated on the held-out test set.

For non-zero-shot evaluation, we create a neuron-wise stratified test-train split with a test size of 20\%. The validation set for finding the best epoch is created from the train fold with a size of 20\%.

\subsection{Steinmetz et al.}

The Steinmetz et al.~dataset consists of 39 Neuropixels recordings from 10 unique mice, with each recording comprising 2 or 3 probe insertions.
For non-zero-shot evaluation, we create train–test splits following the same procedure as described for the IBL dataset (\cref{sec:ibl_data_detail}).

For transductive zero-shot evaluation, we use a 10-fold leave-one-subject-out strategy.
In each fold, one subject is held out for testing, and the remaining nine are used for training.
We select the best training epoch by performing a stratified 80/20 train–validation split within the training set.
For inductive zero-shot evaluation, we hold out 3 subjects for testing and pretrain on the remaining 7.
To select the best classifier epoch, we perform 4-fold subject-wise cross-validation within the training subjects.

\subsection{Bugeon et al.}
\label{sec:bugeon_data_detail}

The Bugeon et al.~dataset contains 17 spatial transcriptomic calcium imaging recordings from 4 unique mice.
We train on all stimulus-specific sub-recordings within these sessions.
Since this is the only optophysiology dataset in our evaluation, we adjusted several hyperparameters specifically for calcium activity.
During development, we held out 4 recordings—one from each subject—as a validation set for tuning model hyperparameters.
These recordings are never included in the test set when reporting final performance.

For non-zero-shot evaluation, we create neuron-wise train–test splits following the same protocol as in the IBL dataset (\cref{sec:ibl_data_detail}).
For transductive zero-shot evaluation, we adopt a 4-fold leave-one-subject-out strategy.
In each fold, we perform 3-fold subject-wise cross-validation within the training subjects to select the best epoch.
For inductive zero-shot evaluation, we hold out one subject (SB028) for testing and pretrain on the remaining three.
As in the transductive case, we perform 3-fold subject-wise cross-validation on the training set to select the best model epoch.

%% file: appendix/baselines.tex
\section{Implementation details for baseline models}
\label{sec:baseline_implementation}

\subsection{NeuPRINT}
\paragraph{Calcium Imaging (Bugeon et al.).}
For the Bugeon dataset, we used the publicly available NeuPRINT implementation and followed a transductive evaluation setup consistent with the original codebase.
The model was first pretrained on the full dataset. For evaluation, the embedding table was reinitialized and optimized again using the self-supervised loss, while keeping the backbone encoder frozen.
The resulting neuron embeddings were then used for downstream classification.

\paragraph{Electrophysiology Datasets.}
For electrophysiology datasets, we modified NeuPRINT’s training loop and epoch structure to accommodate the larger scale and greater diversity of these datasets.
In the original implementation, there is no true notion of an epoch—batches are drawn from individual sessions without ensuring full dataset coverage, and the sampling strategy is manually biased toward neurons with labeled cell types.
This setup does not translate well to larger datasets such as Steinmetz et. al., IBL, and Allen VC.

We instead adopted the standard definition of an epoch, which is a full pass through all training data, and trained for 300 epochs.
We also replaced the session-specific sampling strategy in the original implementation with uniform sampling across all sessions to ensure unbiased data coverage.

Furthermore, because these electrophysiology datasets are substantially larger than Bugeon et al. we did not re-initialize and relearn the embedding table during evaluation. The original NeuPRINT protocol is tailored for the Bugeon et al. dataset, where such re-initialization is computationally feasible. However, for large-scale datasets like Steinmetz, IBL, and Allen VC, this procedure becomes too computationally expensive. Therefore, we adopted a more scalable and simpler protocol that just uses learned embeddings from pretraining for downstream evaluation.

All hyperparameters were retained from the original NeuPRINT setup, except for the learning rate, which we set to $1 \times 10^{-2}$, and the backward context window, which was set to 10.

NeuPRINT relies on behavior features being provided along with the mean and standard deviations for the population activity. The behavioral features used for electrophysiology dataset were as follows:
\begin{itemize}[leftmargin=5mm]
\item \textbf{Steinmetz et al.}: face motion, pupil area, and wheel velocity
\item \textbf{IBL}: wheel velocity
\item \textbf{Allen VC}: running speed
\end{itemize}

\subsection{NEMO}
We evaluated NEMO using the official implementation provided by the original authors upon request.
When adapting the method to new datasets, we retained all hyperparameters from the original paper, with the exception of the waveform template dimensionality, which we adjusted to match the characteristics of each dataset.

\subsection{LOLCAT} We evaluated LOLCAT using the official implementation provided by the original authors upon request. For all experiments (unless otherwise stated below), we used a batch size of 64, a learning rate of $1 \times 10^{-3}$, a weight decay of  $1 \times 10^{-5}$, and a dropout rate of 0.5. The MLP hidden dimensions were set to [64, 32, 16], and 4 attention heads were used. The snippet dropout rate during training was 0.45. We trained the models for 300 epochs with the optimizer outlined in the paper. These default parameters were chosen based on manual exploration of the “reduced” hyperparameter search range and explicit prescriptions in the LOLCAT paper.

\begin{itemize}[leftmargin=5mm]
\item \textbf{Allen VC}: We used the same final hyperparameter selection as the paper with new results on our split of the data.
\item \textbf{IBL}: We used the same hyperparameters as Allen VC.
\item \textbf{Steinmetz transductive}: Batch size was increased to 1024, and the epochs were set to 1000.
\item \textbf{Steinmetz inductive}: The minimum factor was set to 0.01, and we initialized the classes undersampling factors to [8 (HPF), 8 (MB), 0.01 (TH), 0.01 (VIS)].
\item \textbf{Bugeon subclass}: The classes undersampling factors were initialized to [0.01 (Lamp5), 0.01 (Pvalb), 8 (Sncg), 8 (Sst), 8 (Vip)] and the minimum factor to 0.01.
\item \textbf{Bugeon E vs. I}: Class undersampling factors were initialized to [0.01 (E), 8 (I)] and the minimum factor to 0.01.
\end{itemize}

%% file: main.bib
@inproceedings{
yu2025in,
title={In vivo cell-type and brain region classification via multimodal contrastive learning},
author={Han Yu and Hanrui Lyu and YiXun Xu and Charlie Windolf and Eric Kenji Lee and Fan Yang and Andrew M Shelton and Olivier Winter and International Brain Laboratory and Eva L Dyer and Chandramouli Chandrasekaran and Nicholas A. Steinmetz and Liam Paninski and Cole Lincoln Hurwitz},
booktitle={The Thirteenth International Conference on Learning Representations},
year={2025},
url={https://openreview.net/forum?id=10JOlFIPjt}
}

@phdthesis{schneider2025robust,
  doi = {10.7936/KKA8-1A55},
  url = {https://openscholarship.wustl.edu/art_sci_etds/3612},
  author = {Schneider,  Aidan},
  title = {Robust Embeddings of Genetics,  Anatomy,  and State Decoded from a Neuron’s Activity},
  school = {Washington University in St. Louis},
  year = {2025}
}

@article{christensen2022cognition,
  title={Cognition and the single neuron: How cell types construct the dynamic computations of frontal cortex},
  author={Christensen, Amelia J and Ott, Torben and Kepecs, Adam},
  journal={Current Opinion in Neurobiology},
  volume={77},
  pages={102630},
  year={2022},
  publisher={Elsevier}
}

@article{peterson2022learning,
  title={Learning neural decoders without labels using multiple data streams},
  author={Peterson, Steven M and Rao, Rajesh PN and Brunton, Bingni W},
  journal={Journal of Neural Engineering},
  volume={19},
  number={4},
  pages={046032},
  year={2022},
  publisher={IOP Publishing}
}

@inproceedings{urzay2023detecting,
  title={Detecting change points in neural population activity with contrastive metric learning},
  author={Urzay, Carolina and Ahad, Nauman and Azabou, Mehdi and Schneider, Aidan and Atamkuri, Geethika and Hengen, Keith B and Dyer, Eva L},
  booktitle={2023 11th International IEEE/EMBS Conference on Neural Engineering (NER)},
  pages={1--4},
  year={2023},
  organization={IEEE}
}

@article{mcgregor2024failure,
  title={Failure in a population: Tauopathy disrupts homeostatic set-points in emergent dynamics despite stability in the constituent neurons},
  author={McGregor, James N and Farris, Clayton A and Ensley, Sahara and Schneider, Aidan and Fosque, Leandro J and Wang, Chao and Tilden, Elizabeth I and Liu, Yuqi and Tu, Jianhong and Elmore, Halla and others},
  journal={Neuron},
  volume={112},
  number={21},
  pages={3567--3584},
  year={2024},
  publisher={Elsevier}
}

@inproceedings{knight2025unified,
  title={Unified Pretraining on Mixed Optophysiology and Electrophysiology Data Across Brain Regions},
  author={Knight, Ian Jarratt and Arora, Vinam and Azabou, Mehdi and Dyer, Eva L},
  booktitle={Foundation Models for the Brain and Body Workshop --- NeurIPS 2025},
  year={2025}
}

@article{siegle2021survey,
  title={Survey of spiking in the mouse visual system reveals functional hierarchy},
  author={Siegle, Joshua H and Jia, Xiaoxuan and Durand, S{\'e}verine and Gale, Sam and Bennett, Corbett and Graddis, Nile and Heller, Greggory and Ramirez, Tamina K and Choi, Hannah and Luviano, Jennifer A and others},
  journal={Nature},
  volume={592},
  number={7852},
  pages={86--92},
  year={2021},
  publisher={Nature Publishing Group UK London}
}

@article{pak2024distinctive,
  title={Distinctive whole-brain cell types predict tissue damage patterns in thirteen neurodegenerative conditions},
  author={Pak, Veronika and Adewale, Quadri and Bzdok, Danilo and Dadar, Mahsa and Zeighami, Yashar and Iturria-Medina, Yasser},
  journal={Elife},
  volume={12},
  pages={RP89368},
  year={2024},
  publisher={eLife Sciences Publications Limited}
}

@article{jha2024disentangling,
  title={Disentangling the roles of distinct cell classes with cell-type dynamical systems},
  author={Jha, Aditi and Gupta, Diksha and Brody, Carlos and Pillow, Jonathan W},
  journal={Advances in Neural Information Processing Systems},
  volume={37},
  pages={33668--33690},
  year={2024}
}

@article{beau2025deep,
  title={A deep-learning strategy to identify cell types across species from high-density extracellular recordings},
  author={Beau, Maxime and Herzfeld, David J. and Naveros, Francisco and Hemelt, Marie E. and D'Agostino, Federico and Oostland, Marlies and S{\'a}nchez-L{\'o}pez, Alvaro and Chung, Young Yoon and Maibach, Michael and Stabb, Hannah N. and others},
  journal={Cell},
  year={2025},
  doi={10.1016/j.cell.2025.01.041},
}

@article{international2023brain,
  title={A brain-wide map of neural activity during complex behaviour},
  author={Angelaki, Dora and Benson, Brandon and Benson, Julius and Birman, Daniel and Bonacchi, Niccol{\`o} and Bougrova, Kc{\'e}nia and Bruijns, Sebastian A and Carandini, Matteo and Catarino, Joana A and others},
  journal={Nature},
  volume={645},
  number={8079},
  pages={177--191},
  year={2025},
  publisher={Nature Publishing Group UK London}
}

@article{lee2024large,
  title={Large-scale neurophysiology and single-cell profiling in human neuroscience},
  author={Lee, Anthony T and Chang, Edward F and Paredes, Mercedes F and Nowakowski, Tomasz J},
  journal={Nature},
  volume={630},
  number={8017},
  pages={587--595},
  year={2024},
  publisher={Nature Publishing Group UK London}
}

@article{juavinett2019chronically,
  title={Chronically implanted Neuropixels probes enable high-yield recordings in freely moving mice},
  author={Juavinett, Ashley L and Bekheet, George and Churchland, Anne K},
  journal={Elife},
  volume={8},
  pages={e47188},
  year={2019},
  publisher={eLife Sciences Publications, Ltd}
}

@article{tasic2018shared,
  title={Shared and distinct transcriptomic cell types across neocortical areas},
  author={Tasic, Bosiljka and Yao, Zizhen and Graybuck, Lucas T and Smith, Kimberly A and Nguyen, Thuc Nghi and Bertagnolli, Darren and Goldy, Jeff and Garren, Emma and Economo, Michael N and Viswanathan, Sarada and others},
  journal={Nature},
  volume={563},
  number={7729},
  pages={72--78},
  year={2018},
  publisher={Nature Publishing Group UK London}
}

@article{zeng2017neuronal,
  title={Neuronal cell-type classification: challenges, opportunities and the path forward},
  author={Zeng, Hongkui and Sanes, Joshua R},
  journal={Nature Reviews Neuroscience},
  volume={18},
  number={9},
  pages={530--546},
  year={2017},
  publisher={Nature Publishing Group UK London}
}

@article{gouwens2019classification,
  title={Classification of electrophysiological and morphological neuron types in the mouse visual cortex},
  author={Gouwens, Nathan W and Sorensen, Staci A and Berg, Jim and Lee, Changkyu and Jarsky, Tim and Ting, Jonathan and Sunkin, Susan M and Feng, David and Anastassiou, Costas A and Barkan, Eliza and others},
  journal={Nature neuroscience},
  volume={22},
  number={7},
  pages={1182--1195},
  year={2019},
  publisher={Nature Publishing Group US New York}
}

@article{mosher2020cellular,
  title={Cellular classes in the human brain revealed in vivo by heartbeat-related modulation of the extracellular action potential waveform},
  author={Mosher, Clayton P and Wei, Yina and Kami{\'n}ski, Jan and Nandi, Anirban and Mamelak, Adam N and Anastassiou, Costas A and Rutishauser, Ueli},
  journal={Cell reports},
  volume={30},
  number={10},
  pages={3536--3551},
  year={2020},
  publisher={Elsevier}
}

@inproceedings{saeed2021contrastive,
  title={Contrastive learning of general-purpose audio representations},
  author={Saeed, Aaqib and Grangier, David and Zeghidour, Neil},
  booktitle={ICASSP 2021-2021 IEEE International Conference on Acoustics, Speech and Signal Processing (ICASSP)},
  pages={3875--3879},
  year={2021},
  organization={IEEE}
}

@article{zhao2025distinct,
  title={Distinct roles of neuronal phenotypes during neurofeedback adaptation},
  author={Zhao, Yi and Stealey, Hannah M and Lu, Hung-Yun and Contreras-Hernandez, Enrique and Chang, Yin-Jui and Tobler, Philippe and Santacruz, Samantha R},
  journal={bioRxiv},
  pages={2025--05},
  year={2025},
  publisher={Cold Spring Harbor Laboratory}
}

@article{hubel1959receptive,
  title={Receptive fields of single neurones in the cat's striate cortex},
  author={Hubel, David H and Wiesel, Torsten N},
  journal={The Journal of physiology},
  volume={148},
  number={3},
  pages={574},
  year={1959}
}

@article{schneider2025inhibitory,
  title={Inhibitory specificity from a connectomic census of mouse visual cortex},
  author={Schneider-Mizell, Casey M and Bodor, Agnes L and Brittain, Derrick and Buchanan, JoAnn and Bumbarger, Daniel J and Elabbady, Leila and Gamlin, Clare and Kapner, Daniel and Kinn, Sam and Mahalingam, Gayathri and others},
  journal={Nature},
  volume={640},
  number={8058},
  pages={448--458},
  year={2025},
  publisher={Nature Publishing Group UK London}
}

@article{ahad2022learning,
  author={Ahad, Nauman and Dyer, Eva L. and Hengen, Keith B. and Xie, Yao and Davenport, Mark A.},
  journal={IEEE Transactions on Signal Processing}, 
  title={Learning Sinkhorn Divergences for Change Point Detection}, 
  year={2025},
  volume={73},
  number={},
  pages={4347-4363},
  doi={10.1109/TSP.2025.3609697}}

@inproceedings{gao2021simcse,
  title={SimCSE: Simple Contrastive Learning of Sentence Embeddings},
  author={Gao, Tianyu and Yao, Xingcheng and Chen, Danqi},
  booktitle={Proceedings of the 2021 Conference on Empirical Methods in Natural Language Processing},
  pages={6894--6910},
  year={2021}
}

@article{le2020contrastive,
  title={Contrastive representation learning: A framework and review},
  author={Le-Khac, Phuc H and Healy, Graham and Smeaton, Alan F},
  journal={Ieee Access},
  volume={8},
  pages={193907--193934},
  year={2020},
  publisher={IEEE}
}

@article{tolossa2024conserved,
  title={Neurons throughout the brain embed robust signatures of their anatomical location into spike trains},
  author={Tolossa, Gemechu Bekele and Schneider, Aidan M and Dyer, Eva and Hengen, Keith B},
  journal={eLife},
  volume={13},
  pages={RP101506},
  year={2025},
  publisher={eLife Sciences Publications Limited}
}

@article{lee2024physmap,
  title={PhysMAP-interpretable in vivo neuronal cell type identification using multi-modal analysis of electrophysiological data},
  author={Lee, Eric Kenji and G{\"u}l, As{\i}m Emre and Heller, Greggory and Lakunina, Anna and Jaramillo, Santiago and Przytycki, Pawel F and Chandrasekaran, Chandramouli},
  journal={BioRxiv},
  pages={2024--02},
  year={2024},
  publisher={Cold Spring Harbor Laboratory}
}

@inproceedings{chen2020simple,
  title={A simple framework for contrastive learning of visual representations},
  author={Chen, Ting and Kornblith, Simon and Norouzi, Mohammad and Hinton, Geoffrey},
  booktitle={International conference on machine learning},
  pages={1597--1607},
  year={2020},
  organization={PmLR}
}

@article{azabou2021mine,
  title={Mine your own view: Self-supervised learning through across-sample prediction},
  author={Azabou, Mehdi and Azar, Mohammad Gheshlaghi and Liu, Ran and Lin, Chi-Heng and Johnson, Erik C and Bhaskaran-Nair, Kiran and Dabagia, Max and Avila-Pires, Bernardo and Kitchell, Lindsey and Hengen, Keith B and others},
  journal={arXiv preprint arXiv:2102.10106},
  year={2021}
}

@article{mi2023learning,
  title={Learning time-invariant representations for individual neurons from population dynamics},
  author={Mi, Lu and Le, Trung and He, Tianxing and Shlizerman, Eli and S{\"u}mb{\"u}l, Uygar},
  journal={Advances in Neural Information Processing Systems},
  volume={36},
  year={2023}
}

@inproceedings{wu2025neuron,
  title={Neuron Platonic Intrinsic Representation From Dynamics Using Contrastive Learning},
  author={Wu, Wei and Liao, Can and Deng, Zizhen and Guo, Zhengrui and Wang, Jinzhuo},
  booktitle={The Thirteenth International Conference on Learning Representations},
  year={2025}
}

@inproceedings{
azabou2023unified,
title={A Unified, Scalable Framework for Neural Population Decoding},
author={Mehdi Azabou and Vinam Arora and Venkataramana Ganesh and Ximeng Mao and Santosh B Nachimuthu and Michael Jacob Mendelson and Blake Aaron Richards and Matthew G Perich and Guillaume Lajoie and Eva L Dyer},
booktitle={Thirty-seventh Conference on Neural Information Processing Systems},
year={2023},
url={https://openreview.net/forum?id=sw2Y0sirtM}
}

@article{schneider2023learnable,
  title={Learnable latent embeddings for joint behavioural and neural analysis},
  author={Schneider, Steffen and Lee, Jin Hwa and Mathis, Mackenzie Weygandt},
  journal={Nature},
  volume={617},
  number={7960},
  pages={360--368},
  year={2023},
  publisher={Nature Publishing Group UK London}
}

@article{liu2021drop,
  title={Drop, swap, and generate: A self-supervised approach for generating neural activity},
  author={Liu, Ran and Azabou, Mehdi and Dabagia, Max and Lin, Chi-Heng and Gheshlaghi Azar, Mohammad and Hengen, Keith and Valko, Michal and Dyer, Eva},
  journal={Advances in neural information processing systems},
  volume={34},
  pages={10587--10599},
  year={2021}
}

@article{schneider2023transcriptomic,
  title={Transcriptomic cell type structures in vivo neuronal activity across multiple timescales},
  author={Schneider, Aidan and Azabou, Mehdi and McDougall-Vigier, Louis and Parks, David F and Ensley, Sahara and Bhaskaran-Nair, Kiran and Nowakowski, Tomasz and Dyer, Eva L and Hengen, Keith B},
  journal={Cell reports},
  volume={42},
  number={4},
  year={2023},
  publisher={Elsevier}
}

@inproceedings{radford2021learning,
  title={Learning transferable visual models from natural language supervision},
  author={Radford, Alec and Kim, Jong Wook and Hallacy, Chris and Ramesh, Aditya and Goh, Gabriel and Agarwal, Sandhini and Sastry, Girish and Askell, Amanda and Mishkin, Pamela and Clark, Jack and others},
  booktitle={International conference on machine learning},
  pages={8748--8763},
  year={2021},
  organization={PmLR}
}

@article{su2023roformerenhancedtransformerrotary,
  title={Roformer: Enhanced transformer with rotary position embedding},
  author={Su, Jianlin and Ahmed, Murtadha and Lu, Yu and Pan, Shengfeng and Bo, Wen and Liu, Yunfeng},
  journal={Neurocomputing},
  volume={568},
  pages={127063},
  year={2024},
  publisher={Elsevier}
}

@InProceedings{10.1007/978-3-031-19809-0_38,
author="Yeh, Chun-Hsiao
and Hong, Cheng-Yao
and Hsu, Yen-Chi
and Liu, Tyng-Luh
and Chen, Yubei
and LeCun, Yann",
editor="Avidan, Shai
and Brostow, Gabriel
and Ciss{\'e}, Moustapha
and Farinella, Giovanni Maria
and Hassner, Tal",
title="Decoupled Contrastive Learning",
booktitle="Computer Vision -- ECCV 2022",
year="2022",
publisher="Springer Nature Switzerland",
address="Cham",
pages="668--684",
isbn="978-3-031-19809-0"
}

@inproceedings{neuprint,
 author = {Mi, Lu and Le, Trung and He, Tianxing and Shlizerman, Eli and S\"{u}mb\"{u}l, Uygar},
 booktitle = {Advances in Neural Information Processing Systems},
 editor = {A. Oh and T. Naumann and A. Globerson and K. Saenko and M. Hardt and S. Levine},
 pages = {46007--46026},
 publisher = {Curran Associates, Inc.},
 title = {Learning Time-Invariant Representations for Individual Neurons from Population Dynamics},
 url = {https://proceedings.neurips.cc/paper_files/paper/2023/file/9032e5c9ec394ce768a2fa9bdc56af6c-Paper-Conference.pdf},
 volume = {36},
 year = {2023}
}

@inproceedings{kalantidis2020hard,
 author = {Kalantidis, Yannis and Sariyildiz, Mert Bulent and Pion, Noe and Weinzaepfel, Philippe and Larlus, Diane},
 booktitle = {Advances in Neural Information Processing Systems},
 editor = {H. Larochelle and M. Ranzato and R. Hadsell and M.F. Balcan and H. Lin},
 pages = {21798--21809},
 publisher = {Curran Associates, Inc.},
 title = {Hard Negative Mixing for Contrastive Learning},
 url = {https://proceedings.neurips.cc/paper_files/paper/2020/file/f7cade80b7cc92b991cf4d2806d6bd78-Paper.pdf},
 volume = {33},
 year = {2020}
}

@article{Bugeon2022,
  title = {A transcriptomic axis predicts state modulation of cortical interneurons},
  volume = {607},
  ISSN = {1476-4687},
  url = {http://dx.doi.org/10.1038/s41586-022-04915-7},
  DOI = {10.1038/s41586-022-04915-7},
  number = {7918},
  journal = {Nature},
  publisher = {Springer Science and Business Media LLC},
  author = {Bugeon,  Stéphane and Duffield,  Joshua and Dipoppa,  Mario and Ritoux,  Anne and Prankerd,  Isabelle and Nicoloutsopoulos,  Dimitris and Orme,  David and Shinn,  Maxwell and Peng,  Han and Forrest,  Hamish and Viduolyte,  Aiste and Reddy,  Charu Bai and Isogai,  Yoh and Carandini,  Matteo and Harris,  Kenneth D.},
  year = {2022},
  month = jul,
  pages = {330–338}
}

@article{Steinmetz2019,
  title = {Distributed coding of choice,  action and engagement across the mouse brain},
  volume = {576},
  ISSN = {1476-4687},
  url = {http://dx.doi.org/10.1038/s41586-019-1787-x},
  DOI = {10.1038/s41586-019-1787-x},
  number = {7786},
  journal = {Nature},
  publisher = {Springer Science and Business Media LLC},
  author = {Steinmetz,  Nicholas A. and Zatka-Haas,  Peter and Carandini,  Matteo and Harris,  Kenneth D.},
  year = {2019},
  month = nov,
  pages = {266–273}
}

@article{grill2020bootstrap,
  title={Bootstrap your own latent-a new approach to self-supervised learning},
  author={Grill, Jean-Bastien and Strub, Florian and Altch{\'e}, Florent and Tallec, Corentin and Richemond, Pierre and Buchatskaya, Elena and Doersch, Carl and Avila Pires, Bernardo and Guo, Zhaohan and Gheshlaghi Azar, Mohammad and others},
  journal={Advances in neural information processing systems},
  volume={33},
  pages={21271--21284},
  year={2020}
}

@inproceedings{
burg2024most,
title={Most discriminative stimuli for functional cell type clustering},
author={Max F Burg and Thomas Zenkel and Michaela Vystr{\v{c}}ilov{\'a} and Jonathan Oesterle and Larissa H{\"o}fling and Konstantin Friedrich Willeke and Jan Lause and Sarah M{\"u}ller and Paul G. Fahey and Zhiwei Ding and Kelli Restivo and Shashwat Sridhar and Tim Gollisch and Philipp Berens and Andreas S. Tolias and Thomas Euler and Matthias Bethge and Alexander S Ecker},
booktitle={The Twelfth International Conference on Learning Representations},
year={2024},
url={https://openreview.net/forum?id=9W6KaAcYlr}
}

@article{Walker2019inception,
  title = {Inception loops discover what excites neurons most using deep predictive models},
  volume = {22},
  ISSN = {1546-1726},
  url = {http://dx.doi.org/10.1038/s41593-019-0517-x},
  DOI = {10.1038/s41593-019-0517-x},
  number = {12},
  journal = {Nature Neuroscience},
  publisher = {Springer Science and Business Media LLC},
  author = {Walker,  Edgar Y. and Sinz,  Fabian H. and Cobos,  Erick and Muhammad,  Taliah and Froudarakis,  Emmanouil and Fahey,  Paul G. and Ecker,  Alexander S. and Reimer,  Jacob and Pitkow,  Xaq and Tolias,  Andreas S.},
  year = {2019},
  month = nov,
  pages = {2060–2065}
}

@unpublished{Willeke2023,
  title = {Deep learning-driven characterization of single cell tuning in primate visual area V4 unveils topological organization},
  url = {http://dx.doi.org/10.1101/2023.05.12.540591},
  DOI = {10.1101/2023.05.12.540591},
  publisher = {Cold Spring Harbor Laboratory},
  author = {Willeke,  Konstantin F. and Restivo,  Kelli and Franke,  Katrin and Nix,  Arne F. and Cadena,  Santiago A. and Shinn,  Tori and Nealley,  Cate and Rodriguez,  Gabrielle and Patel,  Saumil and Ecker,  Alexander S. and Sinz,  Fabian H. and Tolias,  Andreas S.},
  note = {bioRxiv 2023.05.12.540591},
  year = {2023},
  month = may 
}

@article{Hfling2024,
  title = {A chromatic feature detector in the retina signals visual context changes},
  volume = {13},
  ISSN = {2050-084X},
  url = {http://dx.doi.org/10.7554/eLife.86860},
  DOI = {10.7554/elife.86860},
  journal = {eLife},
  publisher = {eLife Sciences Publications,  Ltd},
  author = {H\"{o}fling,  Larissa and Szatko,  Klaudia P and Behrens,  Christian and Deng,  Yuyao and Qiu,  Yongrong and Klindt,  David Alexander and Jessen,  Zachary and Schwartz,  Gregory W and Bethge,  Matthias and Berens,  Philipp and Franke,  Katrin and Ecker,  Alexander S and Euler,  Thomas},
  year = {2024},
  month = oct 
}

@unpublished{Tong2023,
  title = {The feature landscape of visual cortex},
  url = {http://dx.doi.org/10.1101/2023.11.03.565500},
  DOI = {10.1101/2023.11.03.565500},
  publisher = {Cold Spring Harbor Laboratory},
  author = {Tong,  Rudi and da Silva,  Ronan and Lin,  Dongyan and Ghosh,  Arna and Wilsenach,  James and Cianfarano,  Erica and Bashivan,  Pouya and Richards,  Blake and Trenholm,  Stuart},
  note = {bioRxiv 2023.11.03.565500},
  year = {2023},
  month = nov 
}

@unpublished{Ustyuzhaninov2022,
  title = {Digital twin reveals combinatorial code of non-linear computations in the mouse primary visual cortex},
  url = {http://dx.doi.org/10.1101/2022.02.10.479884},
  DOI = {10.1101/2022.02.10.479884},
  publisher = {Cold Spring Harbor Laboratory},
  author = {Ustyuzhaninov,  Ivan and Burg,  Max F. and Cadena,  Santiago A. and Fu,  Jiakun and Muhammad,  Taliah and Ponder,  Kayla and Froudarakis,  Emmanouil and Ding,  Zhiwei and Bethge,  Matthias and Tolias,  Andreas S. and Ecker,  Alexander S.},
  note = {bioRxiv 2022.02.10.479884},
  year = {2022},
  month = feb 
}

@inproceedings{
azabou2025multisession,
title={Multi-session, multi-task neural decoding from distinct cell-types and brain regions},
author={Mehdi Azabou and Krystal Xuejing Pan and Vinam Arora and Ian Jarratt Knight and Eva L Dyer and Blake Aaron Richards},
booktitle={The Thirteenth International Conference on Learning Representations},
year={2025},
url={https://openreview.net/forum?id=IuU0wcO0mo}
}

@inproceedings{
zhang2025neural,
title={Neural Encoding and Decoding at Scale},
author={Yizi Zhang and Yanchen Wang and Mehdi Azabou and Alexandre Andre and Zixuan Wang and Hanrui Lyu and International Brain Laboratory and Eva L Dyer and Liam Paninski and Cole Lincoln Hurwitz},
booktitle={Forty-second International Conference on Machine Learning},
year={2025},
url={https://openreview.net/forum?id=vOdz3zhSCj}
}

@inproceedings{
liu2022seeing,
title={Seeing the forest and the tree: Building representations of both individual and collective dynamics with transformers},
author={Ran Liu and Mehdi Azabou and Max Dabagia and Jingyun Xiao and Eva L Dyer},
booktitle={Advances in Neural Information Processing Systems},
editor={Alice H. Oh and Alekh Agarwal and Danielle Belgrave and Kyunghyun Cho},
year={2022},
url={https://openreview.net/forum?id=5aZ8umizItU}
}

@inproceedings{
le2022stndt,
title={{STNDT}: Modeling Neural Population Activity with Spatiotemporal Transformers},
author={Trung Le and Eli Shlizerman},
booktitle={Advances in Neural Information Processing Systems},
editor={Alice H. Oh and Alekh Agarwal and Danielle Belgrave and Kyunghyun Cho},
year={2022},
url={https://openreview.net/forum?id=iUOUnyS6uTf}
}

@inproceedings{
loshchilov2018decoupled,
title={Decoupled Weight Decay Regularization},
author={Ilya Loshchilov and Frank Hutter},
booktitle={International Conference on Learning Representations},
year={2019},
url={https://openreview.net/forum?id=Bkg6RiCqY7},
}

@InProceedings{lee2019set,
    title={Set Transformer: A Framework for Attention-based Permutation-Invariant Neural Networks},
    author={Lee, Juho and Lee, Yoonho and Kim, Jungtaek and Kosiorek, Adam and Choi, Seungjin and Teh, Yee Whye},
    booktitle={Proceedings of the 36th International Conference on Machine Learning},
    pages={3744--3753},
    year={2019}
}

@article{steinmetz2021neuropixels,
  title = {Neuropixels 2.0: A miniaturized high-density probe for stable,  long-term brain recordings},
  volume = {372},
  ISSN = {1095-9203},
  url = {http://dx.doi.org/10.1126/science.abf4588},
  DOI = {10.1126/science.abf4588},
  number = {6539},
  journal = {Science},
  publisher = {American Association for the Advancement of Science (AAAS)},
  author = {Steinmetz,  Nicholas A. and Aydin,  Cagatay and Lebedeva,  Anna and Okun,  Michael and Pachitariu,  Marius and Bauza,  Marius and Beau,  Maxime and Bhagat,  Jai and B\"{o}hm,  Claudia and Broux,  Martijn and Chen,  Susu and Colonell,  Jennifer and others},
  year = {2021},
  month = apr 
}

@Misc{xFormers2022,
  author =       {Benjamin Lefaudeux and Francisco Massa and Diana Liskovich and Wenhan Xiong and Vittorio Caggiano and Sean Naren and Min Xu and Jieru Hu and Marta Tintore and Susan Zhang and Patrick Labatut and Daniel Haziza and Luca Wehrstedt and Jeremy Reizenstein and Grigory Sizov},
  title =        {xFormers: A modular and hackable Transformer modelling library},
  howpublished = {\url{https://github.com/facebookresearch/xformers}},
  year =         {2022}
}
